\documentclass[acmtog]{acmart}
\acmSubmissionID{1717}

\usepackage{booktabs} 
\usepackage{wrapfig}
\usepackage[ruled]{algorithm2e} 

\SetAlFnt{\small}
\SetAlCapFnt{\small}
\SetAlCapNameFnt{\small}
\SetAlCapHSkip{0pt}

\acmJournal{TOG}

\usepackage{xspace}
\usepackage{multirow}
\usepackage{array}
\usepackage{makecell} 

\makeatletter
\DeclareRobustCommand\onedot{\futurelet\@let@token\@onedot}
\def\@onedot{\ifx\@let@token.\else.\null\fi\xspace}

\def\eg{\emph{e.g}\onedot} 
\def\ie{\emph{i.e}\onedot} 
 
\def\etc{\emph{etc}\onedot}

\makeatother

\begin{document}
\setcopyright{cc}
\setcctype{by}
\acmJournal{TOG}
\acmYear{2026} \acmVolume{45} \acmNumber{4} \acmArticle{119}
\acmMonth{7} \acmDOI{10.1145/3811400}
\title{Pixel Cube: Diffusion-based Portrait Video Relighting Through Realistic Lighting Reproduction}

\author{Yufan Zhang}
\orcid{0009-0008-8814-6165}
\affiliation{%
  \institution{George Mason University}
  \city{Fairfax}
  \country{USA}}
\email{yzhang82@gmu.edu}

\author{Yu Ji}
\orcid{0009-0006-6032-7736}
\affiliation{%
  \institution{LightThought LLC}
  \city{Fairfax}
  \country{USA}}
\email{yeauxji@gmail.com}

\author{Ayo Ajiboye}
\orcid{0009-0009-2959-2998}
\affiliation{%
  \institution{George Mason University}
  \city{Fairfax}
  \country{USA}}
\email{aajiboye@gmu.edu}

\author{Rundi Wu}
\orcid{0000-0003-0133-8196}
\affiliation{%
  \institution{Columbia University}
  \city{New York}
  \country{USA}}
\email{rundi.wu@columbia.edu}

\author{Yu Guo}
\orcid{0000-0002-3420-6619}
\affiliation{%
  \institution{George Mason University}
  \city{Fairfax}
  \country{USA}}
\email{tflsguoyu@gmail.com}

\author{Changxi Zheng}
\orcid{0000-0001-9228-1038}
\affiliation{%
  \institution{Columbia University}
  \city{New York}
  \country{USA}}
\email{cxz@cs.columbia.edu}

\author{Jinwei Ye}
\orcid{0000-0001-7780-7943}
\affiliation{%
  \institution{George Mason University}
  \city{Fairfax}
  \country{USA}}
\email{jinweiye@gmu.edu}
\authornote{Corresponding author.}

\renewcommand{\shortauthors}{Zhang et al.}

\begin{abstract}
We present a diffusion-based method for relighting dynamic portrait videos with photorealism and temporal consistency. Our method is fueled by a hybrid training dataset that consists of real-captured and rendered dynamic portrait videos with diverse subject appearances, facial motions, head poses, and known lighting conditions. Specifically, we construct an LED-based lighting system for realistic lighting emulation and high-speed video relighting data acquisition. By leveraging the image priors embedded in pre-trained video diffusion models, and using per-frame high dynamic range (HDR) environment map as lighting control, we train a high-performance generative model for realistic and identity-preserving dynamic portrait video relighting. In addition to the environment map control, our model uses a synthesized background image to enable control on the camera's exposure level and color tone. Our model can produce temporally consistent relit portrait video that looks realistic and harmonious under a provided new environment and faithfully preserve the subject's expression and fine facial features, including skin tone, wrinkles, and facial hair. Our model generalizes well to unseen data, in terms of the subject appearance, motion, and lighting condition. We perform extensive experiments on relighting in-the-wild videos with various environment maps and demonstrate practical applications on portrait photography. Results show that our method achieves state-of-the-art performance in photorealism, lighting harmony, and temporal consistency. Our project page: \url{https://yufanzhang82.github.io/PixelCube/}.
\end{abstract}




\begin{teaserfigure}
  \centering
  \includegraphics[width=0.99\textwidth]{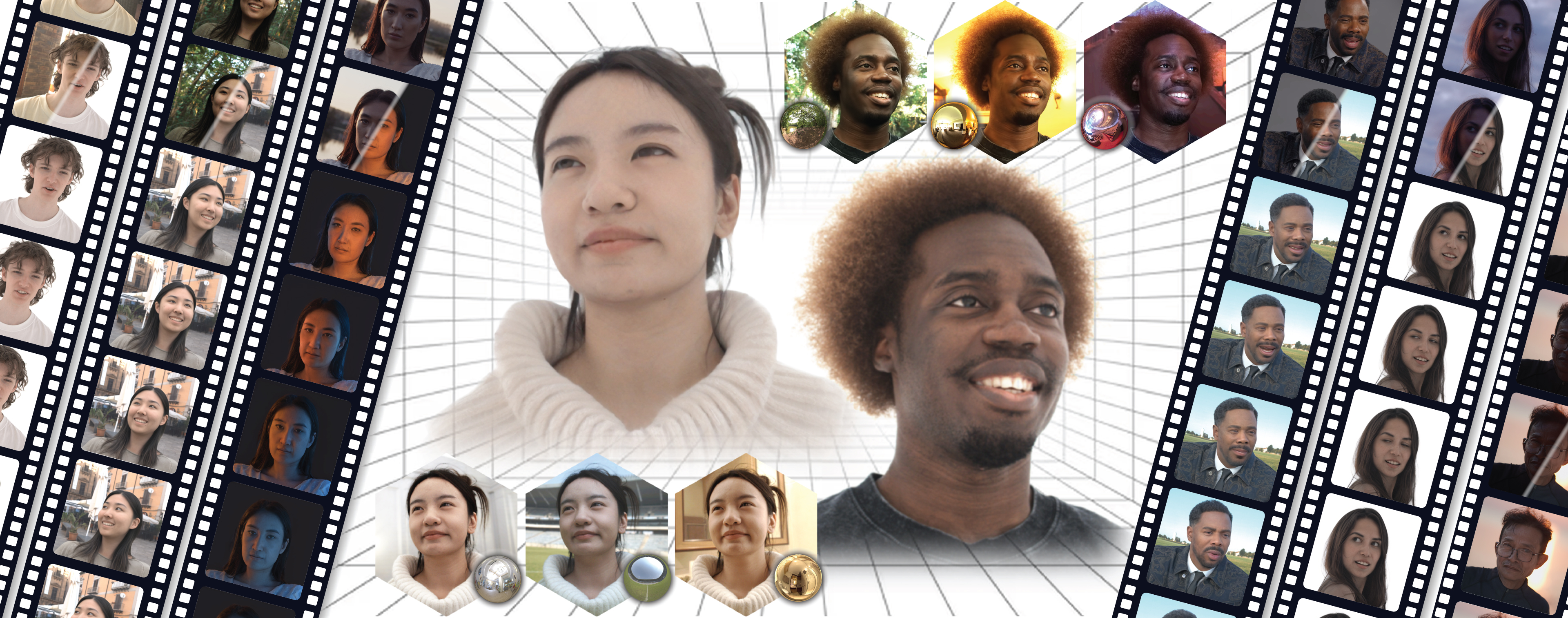}
  \caption{\textbf{We present a diffusion-based model for relighting dynamic portrait videos with photorealism and temporal consistency.} Our model is trained with real-captured and synthetic data, both paired with ground-truth albedo and light maps (see the middle of the figure for examples of our training data). Our model achieves state-of-the-art performance in various portrait relighting applications (left and right).} 
  \Description{teaser figure.}
  \vspace{-2pt}
  \label{fig:teaser}
\end{teaserfigure}


\maketitle

\section{Introduction}
Lighting plays a vital role in producing visually compelling and aesthetically pleasing images and videos. However, in everyday scenarios, optimal lighting conditions are often unavailable, leading to poorly lit portrait subjects. In this work, we introduce a post-capture light-editing method capable of realistically relighting portrait subjects in a captured video to match a provided new environment, whether natural or artificial. 

Synthesizing lighting-consistent portrait images under a new environment is challenging, since the environment light affects many important aspects of the subject's appearance, including shadows, highlights, and colors. 
Even more challenging is relighting videos, as video frames need to be consistent over time.
We present a diffusion-based model trained with high-quality lighting-aware data for relighting dynamic portrait videos with photorealism and temporal consistency. Since high-quality data paired with ground-truth lighting information is critical to fine-tuning the diffusion model, we leverage an LED-based lighting system for training data acquisition. Combining the real-captured footages with rendered ones, we curate a hybrid dynamic portrait video dataset paired with ground-truth light maps, thereby providing rich variations on lighting condition, motion, and subject appearance. 

Most existing learning-based portrait relighting models are trained with portrait images illuminated in a one-light-at-a-time (OLAT) style~\cite{he2024diffrelight,chaturvedi2025synthlight,mei2025lux}. OLAT images are usually captured in a light stage~\cite{debevec2000lightstage1}, equipped with hundreds of individual light sources. The capture process iterates through all the lights. So although OLAT images capture spatially varying lighting conditions, such datasets only provide \emph{static} portrait images. And it remains challenging to adapt OLAT imaging to capture dynamic videos. \citet{Wenger2007TOG} use high-speed synchronized cameras and light sources to cycle through lights fast enough for motion picture acquisition. However, the speed limit of this approach is largely bounded by the data transmission rate and the light source brightness. Even with cameras capturing at 2,000 FPS, the effective video frame rate is only around 15 FPS. Due to this challenge, no existing real-captured dynamic human relighting dataset provides ground-truth light maps~\cite{wang2025comprehensive,mei2025lux}.

\begin{figure}[t]
  \includegraphics[width=\linewidth]{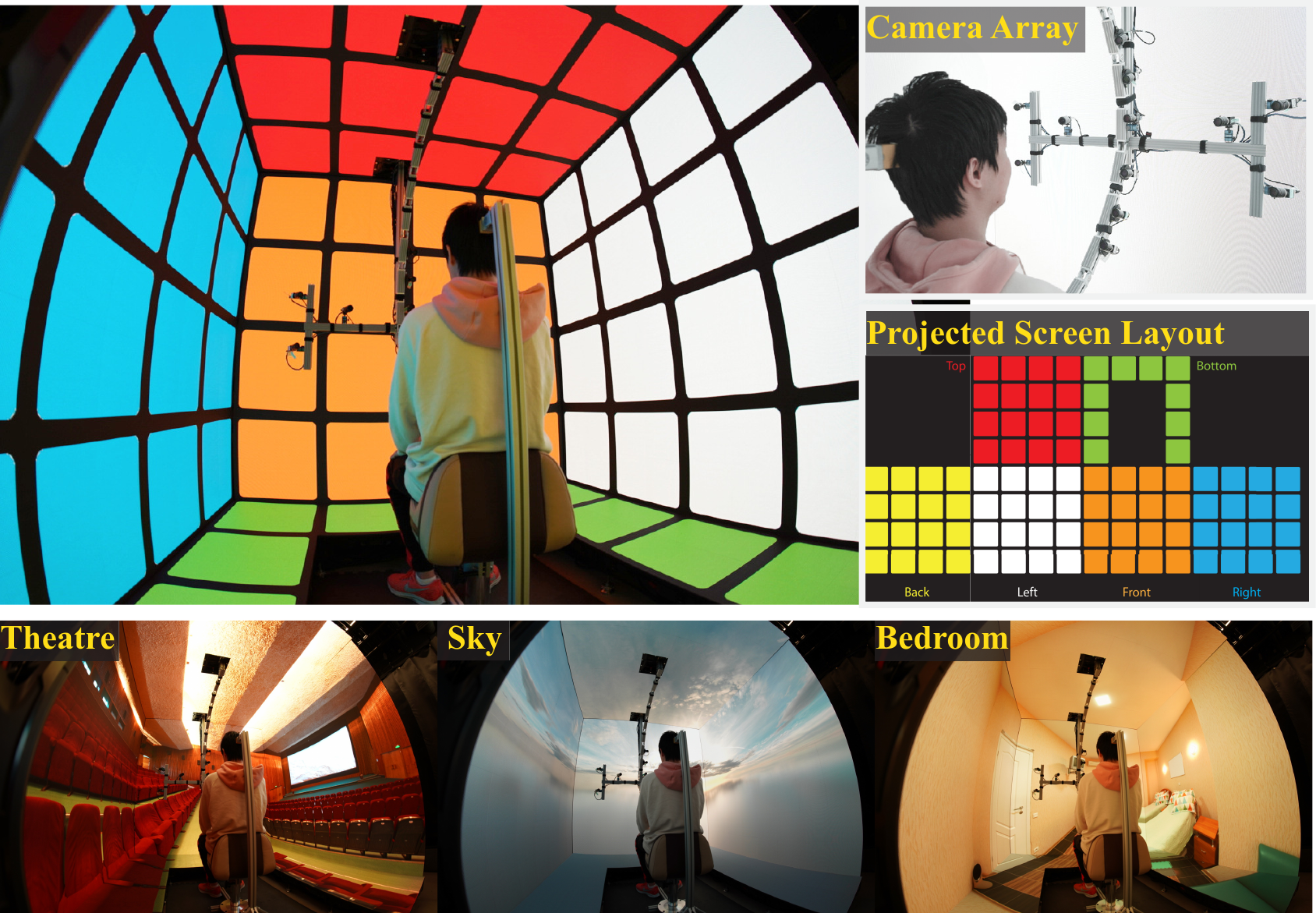}
  \caption{\textbf{\textit{The Pixel Cube}, our lighting system used for training data acquisition.} We show the Pixel Cube with different environment maps (``theater", ``sky", and ``bedroom"), zoom-in view of the acquisition camera array, and the projected panel layout for display.}
  \Description{Imaging system}
  \label{fig:real_sys}
\end{figure}

In this work, we construct the \textit{Pixel Cube} (see Fig.~\ref{fig:real_sys}), a cube-shaped LED stage, for realistic lighting emulation and video relighting data acquisition. LED stages are commonly used as virtual production systems in film making by providing realistic \emph{in-camera background}~\cite{kadner1899wraps,LeGendre2022vp}. In contrast, we use an LED stage with full-surround panels to faithfully reproduce the lighting of a real-world environment by showing its $360^\circ$ HDR environment map. We perform careful radiometric calibration on the system and compensate for the input environment to ensure that our reproduced radiance is consistent with the physical scene radiance\textemdash critical to acquiring relit images as if they were taken in a real physical environment. By synchronizing the acquisition camera with the display refresh rate, our system can capture images lit up by the displayed environment at 60 FPS. We further time-multiplex the display of multiple environment maps to allow capturing the same motion sequence under different environment lights. 
We insert a pure white background in each group of environment maps to capture the target's ground-truth flat-lit albedo.

To supplement real data with more diverse subject appearances (\eg, facial shape, skin tone and texture, hair style, and facial hair), we render dynamic portrait videos of various digital humans, using the MetaHuman plugin of Unreal Engine. MetaHuman provides a set of realistic digital humans that are editable and animatable. With Unreal's high-performance character shader, we render high-quality photorealistic portrait performance videos, paired with ground-truth environment map and per-frame flat-lit albedo. 
 
With our diverse lighting-aware hybrid dataset and leveraging the learned image priors embedded in a pre-trained video diffusion model (\eg, Stable Video Diffusion~\cite{blattmann2023stable}), we fine-tune a generative video model for realistic and identity-preserving portrait video relighting. Our model takes a source portrait video and a target environment map as input. We first delight the source video to estimate its flat-lit albedo, and then relight the albedo using target environment map as light control. Specifically, we encode an environment map to control the lighting via multi-level cross attention. We also use a projected background image to control the exposure level and color tone to further enhance the lighting consistency of relit images. Our model can produce temporally coherent relit videos that appear physically realistic under the provided new lighting condition, while preserving the subject's appearance and motion. Since the video diffusion model is trained with a fixed length, to support long video inference, we infer sequences with overlapping frames. For the overlapped frames, we take the initial noise and per-step clean latents from the previous sequence to guide the inference of the current sequence, in order to enforce across-sequence temporal consistency.
 
We evaluate our relighting model comprehensively. Leveraging the Pixel Cube’s capability to capture identical motion sequences under varied, known illumination, we perform quantitative evaluation. To further demonstrate the robustness of our model, we evaluate its performance on a diverse set of in-the-wild portrait videos and showcase its applications in computational photography and cinematography. Our results indicate that the proposed method achieves state-of-the-art performance, producing photorealistic and temporally coherent relighting while strictly preserving the subject’s identity. 
Furthermore, our model exhibits strong generalization capabilities across unseen subject appearance, complex motion, and novel lighting environment.
  
We summarize our main technical contributions as follows:

\begin{itemize}
 \item We construct the Pixel Cube, an LED-cubic stage for realistic lighting emulation and high-speed relit video acquisition.
 \item We curate a high-quality hybrid (synthetic + real) portrait dataset with ground-truth lighting and flat-lit albedo reference, providing diverse variations on lighting condition, facial and head motion, and subject appearance.
\item We fine tune a video diffusion model for photorealistic, temporally consistent, and identity preserving portrait video relighting. 
\end{itemize}

\section{Related Works}
\paragraph{Portrait Relighting.} Image relighting has been extensively investigated in recent decades, driven by its enormous applications in consumer photography and cinematography. Early works use an OLAT light stage to capture images and 
synthesize relit reflectance by linearly combining those images based on lighting projection~\cite{debevec2000lightstage1}. 
Light stage has evolved to improve the camera configuration~\cite{Debevec2002tog,hawkins2005dual}, light source design~\cite{ghosh2010circularly,ghosh2011multiview}, and illumination scheme~\cite{debevec2004postproduction,sun2020tog}. 
Meanwhile, there exist computational methods that relight portraits 
through geometry reconstruction~\cite{tunwattanapong2013acquiring,fyffe2014driving}, reflectance estimation~\cite{Meka2020tog,debevec2004estimating}, and motion acquisition~\cite{Hawkinsegsr2004,Wenger2007TOG}. 

Relighting can be further achieved through a rendering perspective by combining geometry, material, and lighting information~\cite{1673330}. Some methods treat relighting as an image style transfer problem, transferring the lighting effect from a source subject to a target one, using GAN~\cite{tan2022volux,Ranjan_2023_CVPR}, image decomposition~\cite{li2014intrinsic,hou2022face}, and geometry-aware color transfer~\cite{shih2014style,shu2017portrait}. These methods operate on the image space and are lighting agnostic. As a result, their relit targets may receive lighting effects inconsistent with the background scene. 
Other methods relight portraits by harmonizing the appearance of the foreground subject and the background image~\cite{ren2024relightful,wang2025comprehensive}. 

Recent approaches combine OLAT data with various 3D representations (\eg, NeRF, 3D Gaussians) to alter lighting effects on a volumetric subject~\cite{philip2021free,li2024uravatar,Cai_2024_CVPR,yang2024vrmm,saito2024rgca,schmidt2025becominglit}. These approaches usually require dense multi-view input for 3D volumetric reconstruction. Several convolutional neural networks are trained for relighting using physical-based losses~\cite{pandey2021total,kim2024switchlight}; They typically require geometric priors, such as surface normal. \citet{yeh2022learning} used synthetic data rendered in a virtual light stage for photorealistic portrait relighting. \citet{Mei2024CVPR} integrated a 3D generative model into a relighting network, allowing relighting of a portrait from a free point of view.

\paragraph{Diffusion Models} 
Large latent diffusion models~\cite{rombach2021highresolution,Blattmann2022diffusion}, trained on massive text-image pairs available online, have demonstrated success in generating high-quality images. 
The generative output of a diffusion model can be controlled with text prompts or image conditions, by which diffusion models have been applied in various tasks, such as image editing~\cite{xie2023smartbrush}, image restoration~\cite{xia2023diffir}, depth and normal estimation~\cite{ke2023repurposing,ke2025marigold}, and 3D generation~\cite{poole2022dreamfusion}. 
Those tasks are often achieved by
fine-tuning the model on ground-truth-paired dataset tailored for specific tasks. Further, video diffusion models~\cite{ho2022video,blattmann2023stable} are developed by adding temporal convolution and attention layers in the training pipeline to enforce temporal consistency. Our work repurposes a state-of-the-art video diffusion model, Stable Video Diffusion~\cite{blattmann2023stable}, for dynamic portrait relighting. We fine-tune the model using our high quality portrait video dataset with diverse lighting and subject appearances. We encode an orientation-aware HDR environment map for lighting control.

\paragraph{Diffusion-based Relighting} 
Recent research has leveraged the generative priors of diffusion models to address relighting through various conditioning mechanisms. For static images, IC-Light~\cite{zhang2025scaling} utilizes synthesized OLAT data for prompt-based control, while DiLightNet~\cite{zeng2024dilightnet} and 
LightLab~\cite{lightlab@Magar2025} 
introduce radiance hints and direct light-source manipulation to enhance output quality. 
In the temporal domain, RelightVid~\cite{fang2025relightvid}
facilitates video relighting through multi-modal inputs, including text and environment maps. Furthermore, DiffusionRenderer~\cite{DiffusionRenderer}
incorporates explicit geometric and material properties to condition video diffusion for both forward and inverse rendering. While these methods demonstrate the versatility of diffusion-based lighting control, achieving natural and temporally consistent dynamic portrait relighting remains a significant challenge that requires higher-fidelity training data.

\begin{figure*}[t]
  \includegraphics[width=\linewidth]{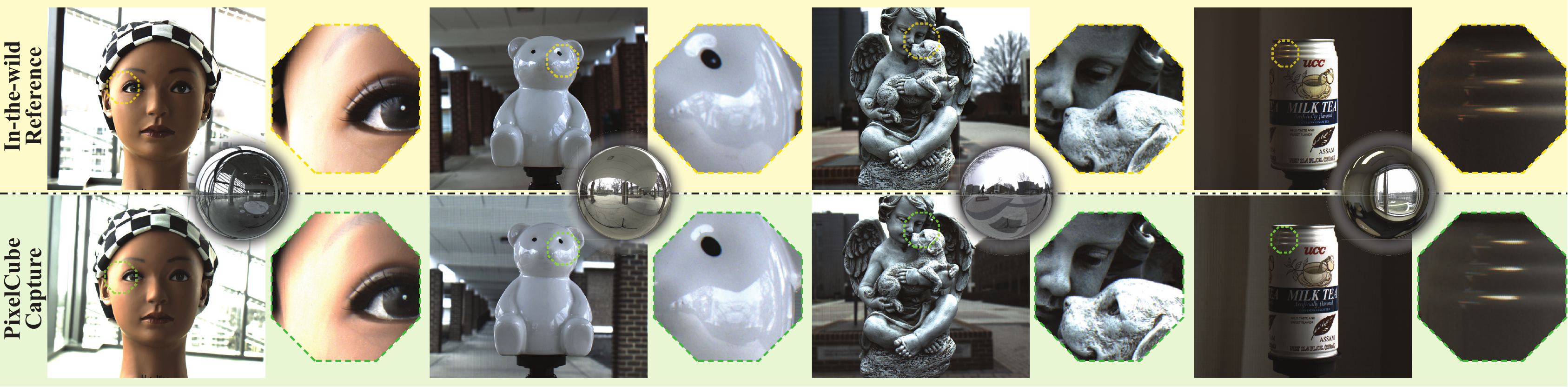}
    \caption{\textbf{Real-world environment lighting emulation.} We compares images taken in the Pixel Cube with those taken under a real-world environment. The spherical environment maps are shown in the center of each group. We display the environment map in the Pixel Cube to emulate the lighting of a real-world environment. We can see that the Pixel Cube can faithfully reproduce the real-world illumination.}
  \Description{Compared with real lighting}
  \label{fig:vs_real}
\end{figure*}
Beyond general-purpose relighting, a specialized branch of research focuses on portrait-specific relighting, where the preservation of intricate facial features\textemdash such as skin texture, wrinkles, and hair\textemdash is paramount. For static images, DiFaReli~\cite{ponglertnapakorn2023difareli} conditions diffusion models on estimated portrait shape and lighting parameters, while SynthLight~\cite{chaturvedi2025synthlight} employs multi-task training on hybrid datasets. 
DiffRelight~\cite{he2024diffrelight} uses multi-view OLAT images to train a diffusion-based portrait relighting model. 
Although this method well preserves the portrait subject's appearance, it is
limited to subject-specific multi-view images taken in their stage. 
Comprehensive Relighting~\cite{wang2025comprehensive}
explores unsupervised learning from in-the-wild videos, and
3DPR~\cite{prao20253dpr} trains a diffusion model with OLAT data for static portrait relighting and novel view synthesis.
Lux Post Facto~\cite{mei2025lux} 
trains their relighting model on a combination of static OLAT data and motion-heavy in-the-wild videos. 
Nevertheless, a persistent challenge in these video-based methods is the absence of ground-truth illumination maps, often necessitating a reliance on appearance-based transfer. In contrast, our framework is trained on a hybrid dataset featuring dynamic portrait videos paired with both ground-truth environment maps and per-frame flat-lit albedo, enabling a more physically accurate and robust fine-tuning process for temporally coherent relighting.

\section{Lighting System for Data Acquisition}
\label{sec:system}
The efficacy of diffusion-based relighting frameworks requires training on diverse, high-fidelity images captured under known illumination. To facilitate training data acquisition, we construct the Pixel Cube (see Fig.~\ref{fig:real_sys}), a cubic LED-based lighting stage designed to emulate complex, full-surround environments. In contrast to traditional light stages, which typically utilize sparse point-light sources and sequential acquisition, the Pixel Cube offers advantages in both temporal resolution and radiometric realism:
1) by synchronizing the cameras with the LED panels' refresh cycle, the system enables the capture of dynamic sequences at 60 FPS; and
2) the high-resolution LED array provides nearly continuous angular sampling, allowing for the faithful reproduction of complex real-world radiance distributions (see Fig.~\ref{fig:vs_real}). 

\paragraph{System Design.}
The Pixel Cube is a 2m $\times$ 2m $\times$ 2m cubic apparatus featuring an interior lined with 90 high-resolution LED panels. Each panel has a resolution of $864 \times 864$ pixels and a peak luminance of 1,900 nits, providing the high dynamic range (HDR) necessary for realistic lighting emulation. To allow high-speed capture, we designed a custom control interface utilizing a tri-level synchronization protocol, which enables stable refresh rates up to 120 FPS. Environmental illumination is reproduced by remapping $360^{\circ}$ panoramic HDR maps into a cube map configuration that corresponds to the interior panel geometry (see Fig. 2). During acquisition, the subject is positioned at the center of the cube so that they are enveloped by physically accurate, full-surround radiance.

For multi-view image acquisition, the system utilizes 12 machine vision cameras, each featuring $2448 \times 2048$ pixels and equipped with 16mm fixed-focal-length lenses. 
The cameras are carefully positioned to capture the subject from diverse viewpoints, with the optical focus centered on the facial region. To maintain temporal alignment between the illumination and capture phases, the cameras operate at 60 FPS, driven by hardware trigger signals synchronized with the LED panels' refresh cycle. We perform geometric and radiometric calibration to establish precise extrinsic and intrinsic parameters while ensuring chromatic consistency across the entire camera array.

\paragraph{Radiometric Compensation.}
To ensure the Pixel Cube faithfully reproduces a real-world lighting condition from a given HDR environment map, we perform radiometric calibration and then compensate the intensity of the input environment map for realistic lighting emulation. Specifically, we take into account three factors: 1) radiance linearization, 2) angular and distance falloff, and 3) color-dependent attenuation.

Radiance linearization is essential to ensure that the radiance of light reproduced by our system is proportional to the actual scene radiance. In this step, we use raw linear images to calibrate the display response curve~\cite{Debevec97sig} of the LED panel used in our cube. Specifically, we take images of a small, uniform patch of the LED panel using the frontal parallel camera with a zoom lens. We set the camera slightly out-of-focus to suppress the Moir\`{e} patterns. We sweep all possible intensity values to establish to a mapping between input intensity and emitted radiance. The response curve of each color channel is measured separately. According to our measurements, we use gamma functions to proximate the display response curves. We then linearize the emitted radiance by applying the inverse gamma function to compensate the intensity of the input environment map: 

\begin{equation}
I_c = \left( \frac{L_c}{L_{\max}} \right)^{\frac{1}{\gamma}},
\label{eq:display_gamma_correction}
\end{equation}
where $I_c$ is the input image intensity with subscript $c \in \{R, G, B\}$ indicating the color channel; $L_c$ denotes the actual radiance of the environment light; $L_{\max}$ is the maximum displayable radiance of our LED panel; and $\gamma$ is the fitted exponent of the display response curve.

\begin{figure}[t]
  \includegraphics[width=\linewidth]{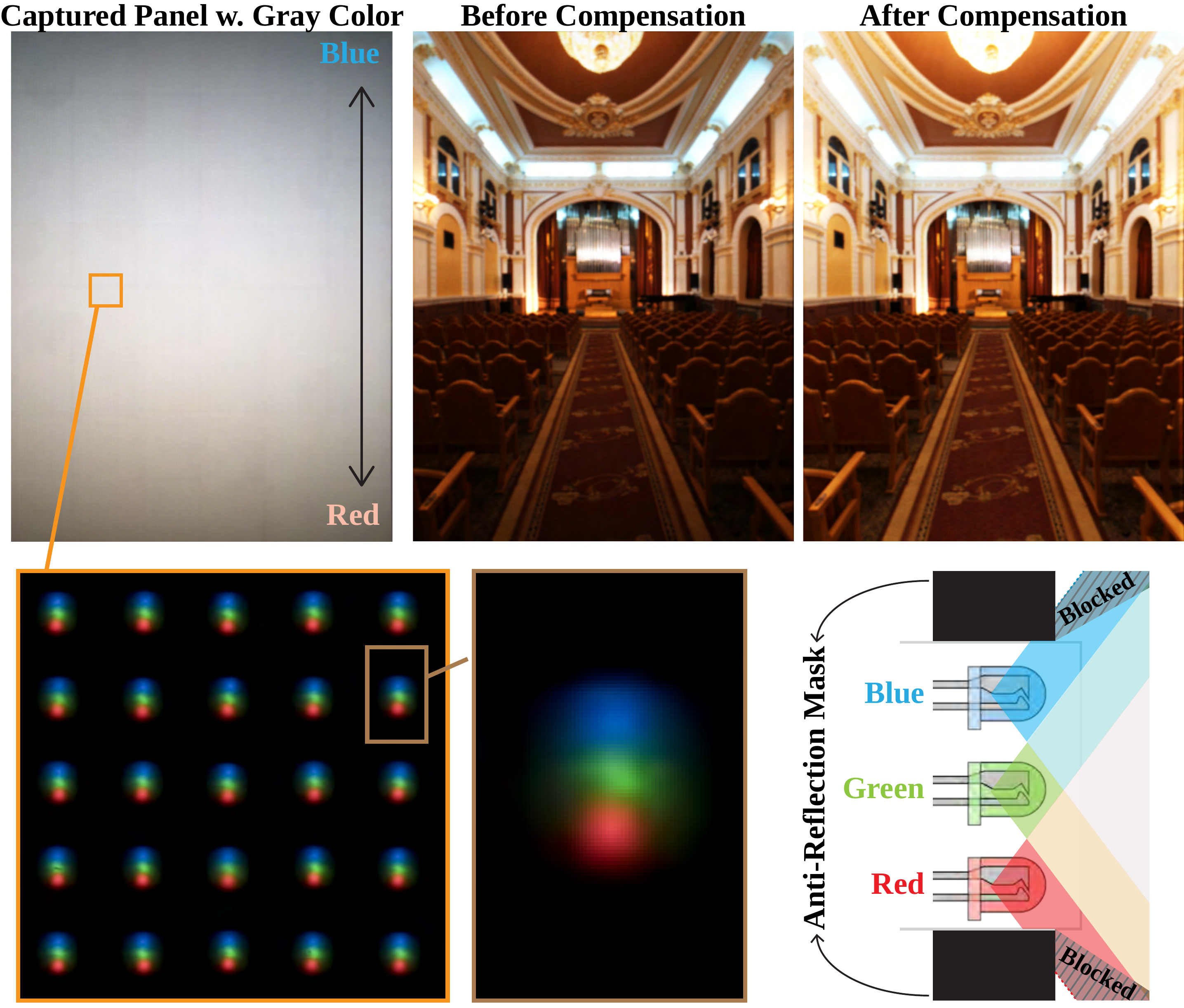}
  \caption{\textbf{Radiometric compensation.} (a) One side of the Pixel Cube wall with gray color displayed (without compensation); (b) Zoom-in views of pixels on the LED panel; (c) Illustration of the anti-reflection mask that blocks peripheral light; (d) Comparison of an image displayed on the cube wall before vs. after color and intensity compensation.}
  \Description{Color compensation}
  \label{fig:color_correction}
\end{figure}

Because of the cube's planar surface structure, we need to take into account the distance and angular falloff of emitted radiance. Assuming the target is at the center of the cube, given the size of our cube ($2$m $\times$ $2$m $\times$ $2$m), the radiance received from a corner pixel of the cube would suffer from 50\% of attenuation comparing to a pixel at the center of a face, due to the distance and angular falloff. This would result in the peripheral pixels on each side of the cube appearing darker than the center ones, which is similar to the vignetting artifact in images. 

Besides this uniform falloff, we also observe color-dependent attenuation caused by the anti-reflection mask. This black mask is applied on the panel surrounding each pixel to absorb light and reduce undesired reflection. However, they also slightly block the peripheral light emitted from a LED photodiode, as shown in Fig.~\ref{fig:color_correction} (c). Since the mask that covers around each pixel is composed of three vertically stacked photodiodes, this occlusion results in color-dependent intensity attenuation that leads to color shifting along the vertical direction. In particular, as shown in Fig.~\ref{fig:color_correction}, when showing a neutral color, the upper side of a vertical face appears bluish while the lower side appears reddish.

We jointly compensate for the distance/angular fall off and color-dependent attenuation. To calibrate the light attenuation, we display a neutral image (\ie, $R = G = B$) and capture it using a wide-angle camera from the center of the cube (we do this for all cube faces). We then warp the captured image to the LED panel layout and calculate per-channel attenuation coefficients. We multiply the inverse of these coefficients with the input image to compensate for the color-dependent attenuation. This compensation is applied after gamma correction. An example of displayed image before vs. after this compensation is shown in Fig.~\ref{fig:color_correction}. We can see that the compensation effectively corrects the intensity fall off and color shifting.

\paragraph{Real-world Environment Emulation}
Here we show that the Pixel Cube is able to faithfully reproduce the lighting effect of a real-world environment. Specifically, we perform experiments by taking pictures of various objects with different reflectance properties under a real-world environment and in the Pixel Cube with replicated lighting.

\begin{wrapfigure}{r}{0.35\linewidth} 
\vspace{-6pt}
    \centering
    \includegraphics[width=0.35\columnwidth]{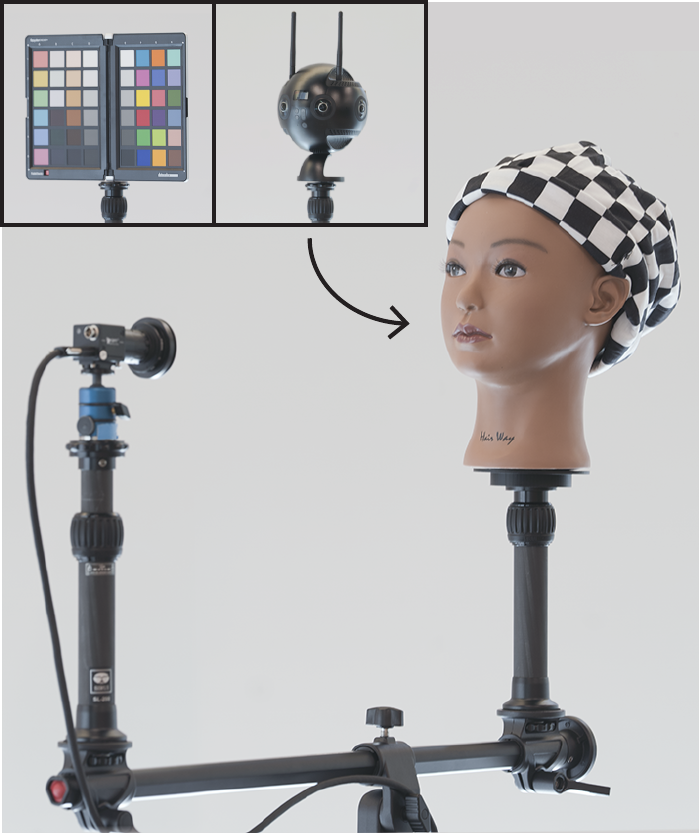} 
    \caption{Setup for lighting comparison experiments.}
    \vspace{-8pt}
    \label{fig:real_lighting}
\end{wrapfigure}
Fig.~\ref{fig:real_lighting} shows our experimental setup. We mount the camera and the target on a rigid rig, so the two images taken under the real-world environment and in the Pixel Cube are from the same viewpoint. The target can be swapped to other objects. We experiment with four objects with different levels of specularity. Before taking the target pictures, we always capture an image of a color checker in order to unify the color tone of captured target images. For each object, we test in four environments, all with highly contrastive lighting. To acquire the environment map, we put a $360^\circ$ camera in place of the target to take panoramic images of the environment. We use five different exposure levels to acquire HDR scene radiance. The HDR environment map is then linearized and color compensated for showing in our cube. 

When imaging in the cube, we put the imaging rig in the center of the cube and orient the view direction, such that the target orientation w.r.t. the environment matches the setting of the real-world capture. Our lighting emulation results are shown in Fig.~\ref{fig:vs_real}. We can see that targets' appearances under the two environments (real-world environment vs. environment replicated by the Pixel Cube) closely resemble each other in terms of shadows and specular highlights. For example, the specular reflections on the ceramic bear are almost identical. These comparisons verify the Pixel Cube's ability on realistic real-world environment lighting reproduction.  

\begin{figure}[t]
  \includegraphics[width=0.87\linewidth]{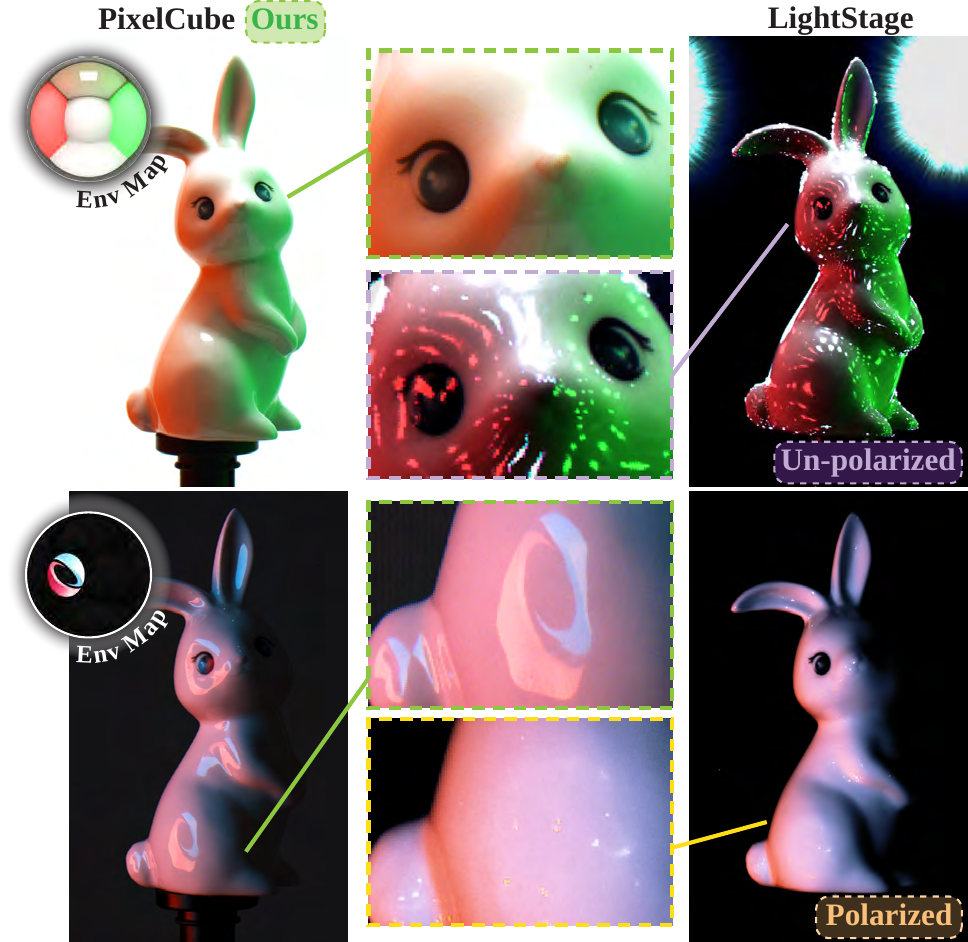}
  \caption{\textbf{Lighting emulation comparison between the Pixel Cube and the light stage.} Here we compare the lighting effect reproduced by the Pixel Cube and the light stage, given an environment map. We compare against two versions of light stage: standard imaging (top row) and polarized imaging (bottom row). }
  \Description{Compared with Light Stage}
  \label{fig:vs_lightstage}
\end{figure}

\paragraph{Pixel Cube vs. Light Stage.} Here we compare our Pixel Cube with the light stage in terms of lighting effect reproduction. When compare the images of the same object taken in these two systems when given the same environment map. 

The light stage we use has 148 LED light units uniformly mounted on a geodesic sphere with diameter ~$1m$. Each light unit has six LED photodiodes arranged on the vertices of a hexagon. All LEDs are covered with linear polarizers. On the imaging side, we capture images both by using and not using a linear polarizer. We refer to the two modes as standard imaging (without using a polarizer) and polarized imaging (which uses a linear polarizer).
To simulate the lighting effect of a given environment map, we capture images of the target using the One-Light-At-a-Time (OLAT) scheme, and simulate the lighting effect using the weighted combination of OLAT images. In contrast, in the Pixel Cube, the relit images are taken in one shot by displaying the warped environment map. 

Fig.~\ref{fig:vs_lightstage} shows the light emulation results using two environment maps. The standard imaging mode of light stage preserves specular reflection on objects, while the polarized imaging mode suppresses reflection by using the linear polarization. In the standard mode images, we can see scattered specular highlights, which are resulted by the sparse distribution of light. In contrast, the lighting effect emulated by our Pixel Cube is smooth and continuous since the LED panels emit a dense field of light. In the light stage's polarized imaging mode, although the scattered specular reflections are largely eliminated, resulting in a much smoother appearance of the object, it alters the object's reflectance property and is thus not suitable for relighting applications (although this is desirable for geometry estimation). In sum, the Pixel Cube offers clear advantages over the light stage in both acquisition speed and lighting emulation. 

section{Training Dataset Curation}
Fine-tuning a diffusion model for portrait relighting requires a high-quality, diverse dataset of portraits with ground-truth lighting information. However, curating such a dataset for dynamic portrait video relighting presents significant challenges. Existing OLAT datasets provide ground-truth lighting only for static portraits, 
and it is unclear how easily the light stage can be extended for real-time dynamic video capture. Although a large volume of dynamic portrait videos is readily available online, these resources lack the necessary ground-truth lighting annotations, limiting their utility for supervised training.

To this end, we curate a hybrid portrait video dataset, which consists of real-captured and rendered data, both with ground-truth lighting information. Let us denote a video entry in our dataset as $\mathbf{V}\in\mathbb{R}^{f\times w\times h \times 3}$ (where $f$ is the total frame number, $w$ is the video's width, and $h$ is the video's height). Each $\mathbf{V}$ is paired with its ground-truth HDR environment map $\mathbf{E}\in\mathbb{R}^{512\times 512 \times 3}$, background image $\mathbf{B} \in\mathbb{R}^{ w\times h\times 3}$, per-frame matting mask $\mathbf{M}\in\mathbb{R}^{f\times w\times h}$ and per-frame flat-lit albedo reference $\mathbf{A} \in\mathbb{R}^{f\times w\times h \times 3}$:
\begin{equation}
\mathbf{V}\leftrightarrow\{\mathbf{E}, \mathbf{B},  \mathbf{M},  \mathbf{A}\}.
\end{equation}

On the one hand, real-captured data ensures the photorealism of the diffusion output. On the other hand, the rendered data enriches the variations of subject appearance and facial motion of real data. With the flat-lit albedo reference, we are able to train a diffusion model for canceling out the portrait subject's original lighting effect. We then use the estimated albedo for relighting, which greatly reduces the problem's complexity. In the following, we provide the details on how our dataset is curated and collected. 

\subsection{Real Data Acquisition}
Our real data is captured in the Pixel Cube. We linearize and color compensate an environment map using the method described in Section~\ref{sec:system}. We then display the environment map to emulate the lighting of a real-world environment. We use 12 cameras, synchronized with the display refresh rate, to take videos of a subject, sitting in the center of the cube. The multi-view images enrich our dataset with head pose and viewpoint variations. We develop a Vulkan shader for shuffling the environment maps at a fixed frame rate. By using a time-multiplexed acquisition scheme, we are able to capture identical motions under different lighting conditions.

\begin{figure}[t]
  \includegraphics[width=1\linewidth]{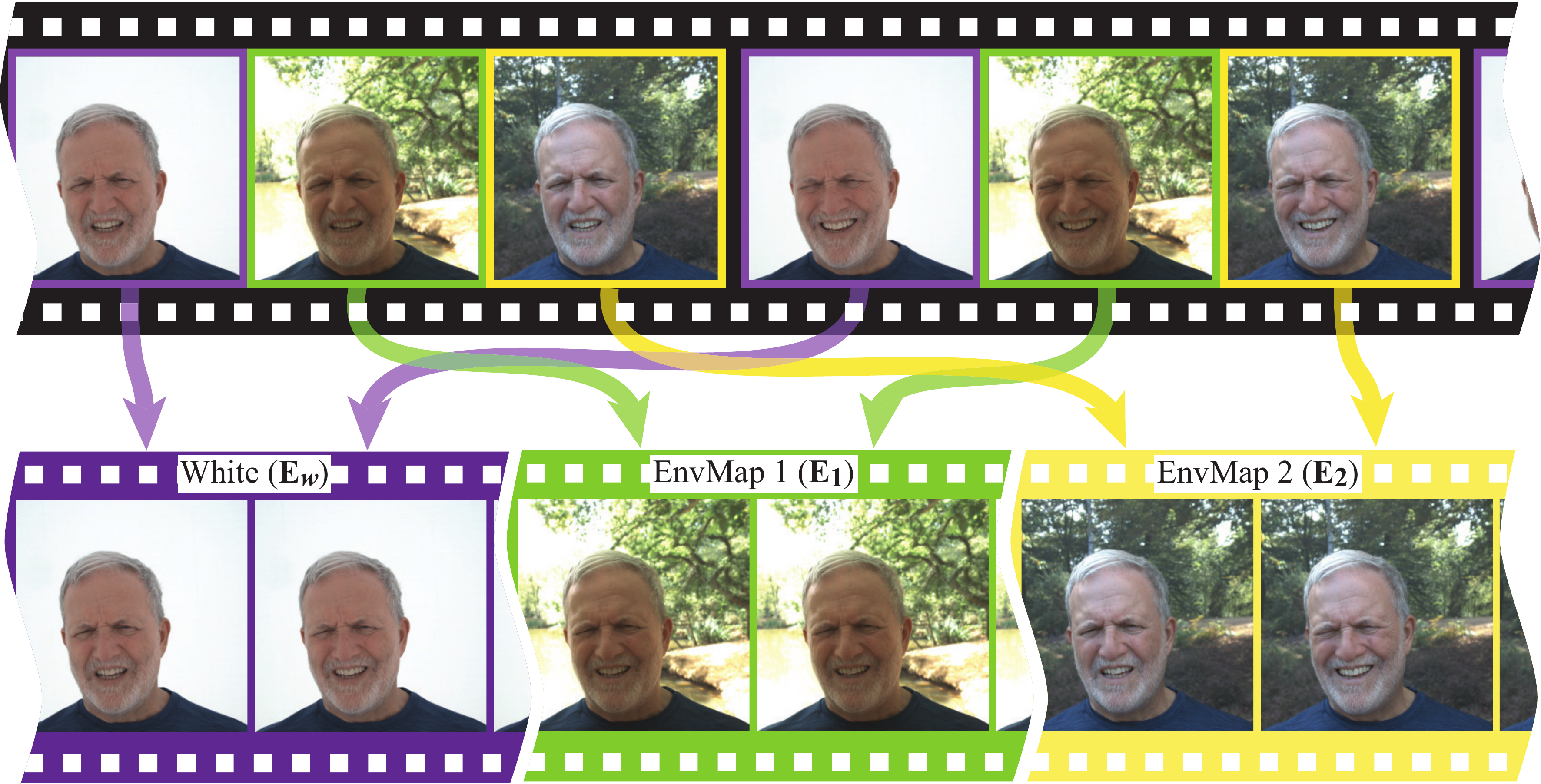}
  \caption{\textbf{Our acquisition scheme.} We interleave a white background and two environment maps for display. By re-arranging the frames, we obtain three sequences with the same motion, but under different lighting.  }
  \Description{acquisition scheme}
  \label{fig:sequence}
\end{figure}

\paragraph{Acquisition Scheme} We adopt a time-multiplexed acquisition scheme that interleaves the display of multiple environment maps to obtain almostly synchronized motion sequences under different lighting conditions. Our acquisition scheme is illustrated in Fig.~\ref{fig:sequence}. Specifically, during acquisition, we interleave a pure white background ($\mathbf{E}_w$) and two different environment maps ($\mathbf{E}_1$ and $\mathbf{E}_2$) for display. We alternating the display of the three environment maps with the following order.: $\mathbf{E}_w \rightarrow\mathbf{E}_1\rightarrow\mathbf{E}_2\rightarrow\mathbf{E}_w \rightarrow ...$. Since our cameras and LED panels are synchronized, we can re-organize the interleaved frame images into three motion sequences, each under the same environment light. Since the camera captures at 60 FPS, the temporal delay between each motion sequence is $16$ ms, which can be neglected when the motion is not drastic. The pure white background $\mathbf{E}_w$ provides us a sequence that can be considered as the ground-truth flat-lit albedo. Therefore, the two other sequences under $\mathbf{E}_1$ and $\mathbf{E}_2$, both paired with albedo reference and ground-truth lighting (\ie, the environment map), can be used for training or evaluating video relighting models.

\begin{figure}[t]
  \includegraphics[width=0.95\linewidth]{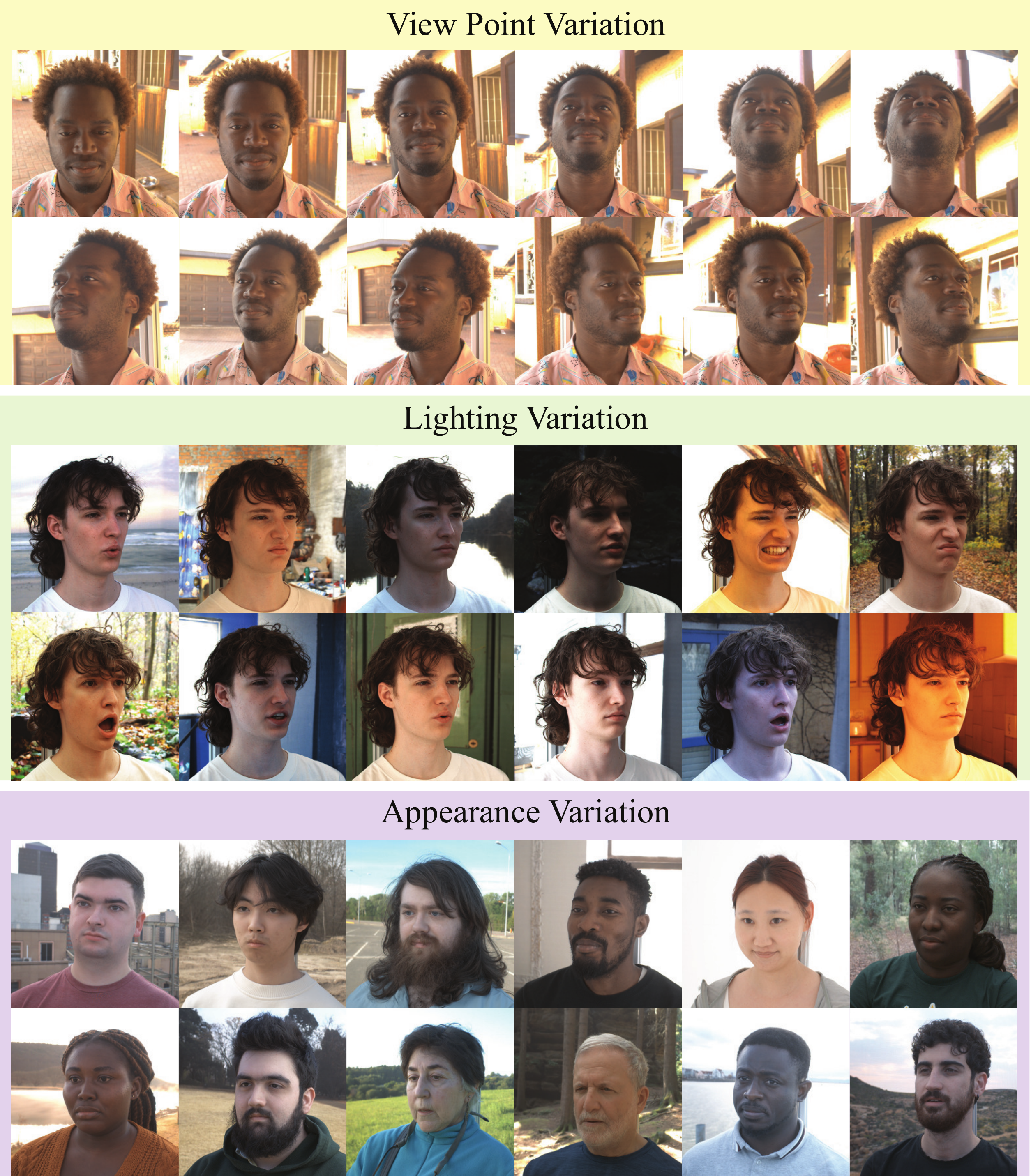}
  \caption{\textbf{Sample images from our real dataset.} We show sample frame images from our real-captured dataset to illustrate its variations in viewpoint, lighting condition, and subject appearance.}
  \label{fig:real data examples}
\end{figure}

\paragraph{Subjects and Environment Light.} We capture portrait videos of 24 subjects (18 males and 6 females), who voluntarily participated in our project and agreed to release their photos for research use. The subjects' ages range from 19 to 75 years old. For each subject, we use the above acquisition scheme to record portrait performance videos under different environment lighting. We sample around 780 HDR environment maps from the Poly Haven HDRI dataset\footnote{https://polyhaven.com} that includes various real-world environments, including indoor, nature, urban, sky light, and studio light. We further rotate the environment map to introduce more lighting variations. In total, we use around 3,000 environment maps for display. We display the environment maps in two modes: static mode and rotating mode. In the rotating mode, we sequentially display a rotating environment map with a $6^\circ$ step. With such data, the model learns to decouple lighting motion and subject motion.

For each subject, we records around 8 video sequences with different facial expressions or natural talking motions. Each video sequence is 12 seconds, which is further re-arranged into two 4-second clips under different environment maps, and a flat-lit albedo reference. In total, our real-captured data has over 2 million frame images, offering diverse variations in lighting, subject appearance, motion, and viewpoint. Fig.~\ref{fig:real data examples} shows example images from our real-captured dataset. 

\subsection{Synthetic Data Generation}
Since the number of subjects in our real-captured data is limited, we use rendered synthetic portrait videos to supplement our dataset in order to introduce more variations in subject appearance, facial motion, and head pose. 

\begin{figure}[t]
  \includegraphics[width=0.9\linewidth]{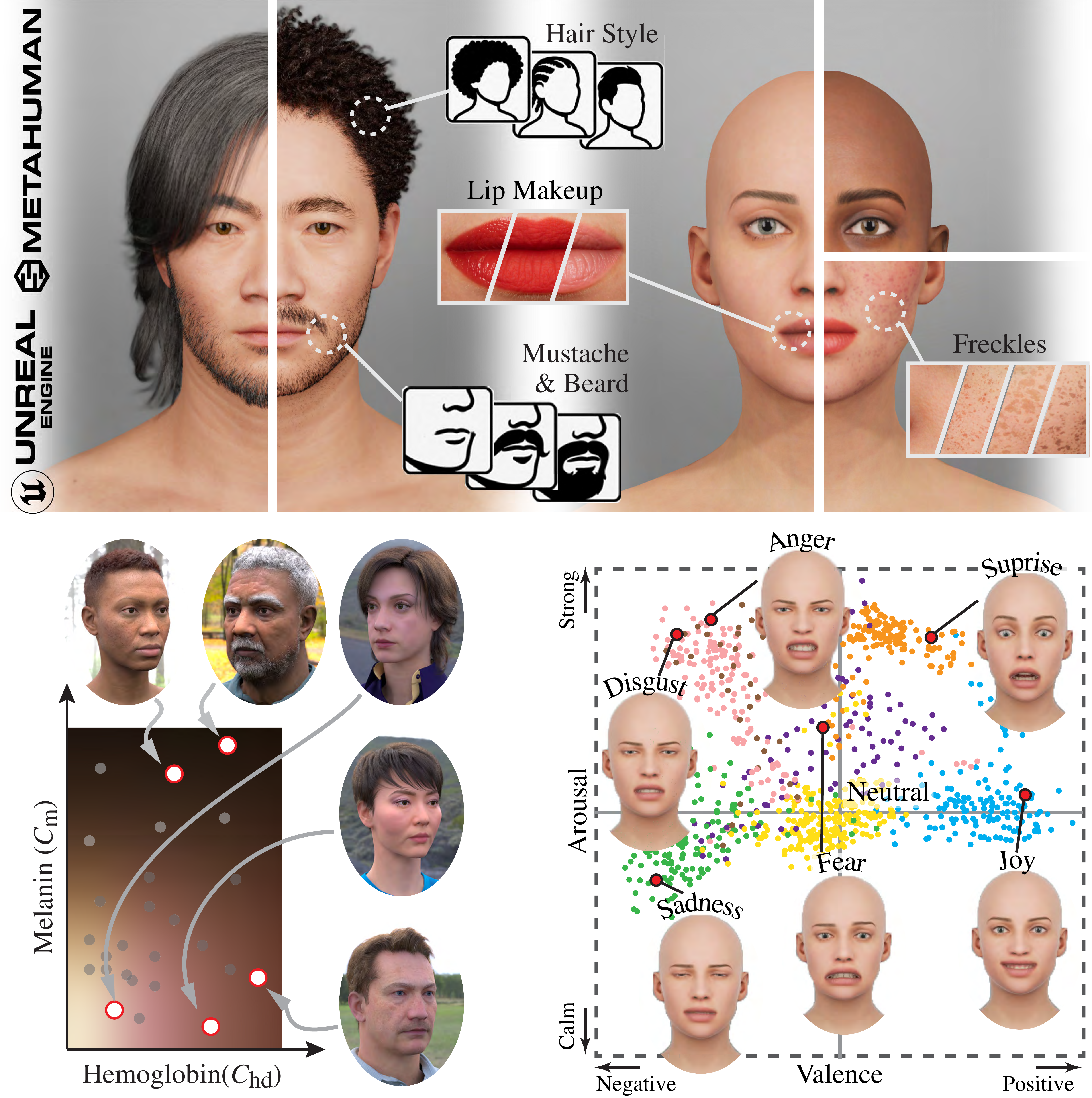}
  \caption{\textbf{Variations of our digital characters.} The digital characters that we use exhibit a large range of variations, in terms of facial features, hair style, skin tone and texture, facial hair, and expression.}
  \Description{Variations of our digital human avatars. }
  \label{fig:synthetic}
\end{figure}

\begin{figure*}[t]
  \includegraphics[width=0.9\linewidth]{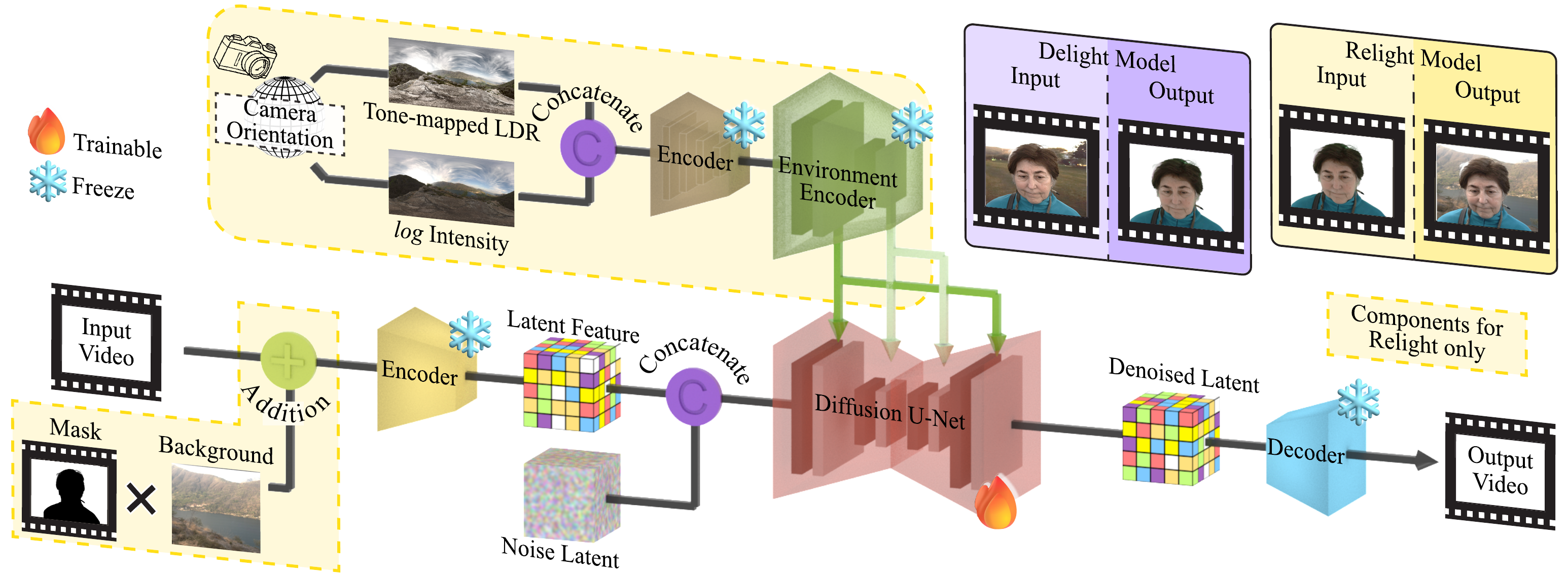}
  \vspace{-10pt}
  \caption{\textbf{The training pipeline of our diffusion models.} Our delight and relight models use the same training backbone, except that the relight model has extra components for lighting conditioning (shown in yellow boxes). }
  \label{fig:pipeline}
\end{figure*}
We use the MetaHuman plugin of Unreal Engine to render the synthetic data. MetaHuman is a powerful framework that allows the creation, animation, and rendering of highly realistic digital human characters. It provides sixty six 3D human presets, scanned from real people, with realistic geometry, skin texture, hair, facial hair, and clothes. The preset characters offer a wide range of facial features, including complexions, skin tones, hair styles, \etc. The framework also provides an interface to allow the users to make plausible adjustments to the preset characters. MetaHuman characters can be animated by directly mirroring the motion of real-captured videos. The framework uses precise keypoint control to map the motion of face, head, and torso. In this way, we use online facial performance videos, as well as selfie videos taken by ourselves to drive the animation of digital characters.

Unreal engine provides a physical-based character shader for rendering high-quality portrait images with photorealistic skin and hair effects. We set up multi-view cameras in the same way as our real system and adopt the same interweaving acquisition scheme (\ie, one white background and two different environment maps per motion sequence) to render the synthetic data. Since we can stop the motion to render different backgrounds, the interleaved motion sequences are perfectly synchronized for the synthetic data. We use the same set of HDR environment maps to provide lighting variations. For each character, we render 3,600 frames with different motions and environment maps. Since we use 30 characters and each are rendered under 12 viewpoints, our synthetic data has around 1.3 million frame images. 

\paragraph{Character Variations.} Our digital characters offer a wide range of variations that help improve the generalization ability of the fine-tuned diffusion model. Fig.~\ref{fig:synthetic} illustrates some of these variations. Our characters are balanced by gender (15 males and 15 females). Each character has distinct facial shapes (face, eyes, and nose), different hairstyles (including the color), facial hair (\eg, eyebrows, mustaches, and beards), and skin tones and textures (\eg, wrinkles and freckles). Some of them wear makeup with different lip colors and glossiness, as well as different eye shadow styles. 

As skin tone is a critical variation across different ethnic groups, we follow~\cite{Jimenez2010TOG} to choose a balanced set of skin colors. Specifically, we use the skin color look-up table parameterized by Melanin and Hemoglobin volume fractions ($C_m$ and $C_{hd}$, respectively) to guide our selection. They can be converted to the $uv$ index used in the MetaHuman skin color picker as: $u=\sqrt[3]{C_m}$, $v=\sqrt[3]{C_{hd}}$. We choose $uv$ indices that result in evenly distributed skin colors in the chart. The distribution of skin colors we use is shown in Fig.~\ref{fig:synthetic}. 

For facial motion, we record videos of facial expressions with a broad spectrum of emotions. We also use online facial performance videos to introduce additional variation. Fig.~\ref{fig:synthetic} visualizes the distribution of our emotional expressions using the arousal-valence model~\cite{toisoul2021estimation} which characterizes the emotion based on the level of intense (arousal) and the positive/negative level (valence). Apart from the expressions, we also use talking faces.

\section{Diffusion Model for Portrait Video Relighting}
We use our high-quality hybrid dataset with ground-truth light map and flat-lit albedo to train a diffusion model for portrait relighting. Fig.~\ref{fig:pipeline} illustrates our training pipeline. We use the pre-trained Stable Video Diffusion (SVD)~\cite{blattmann2023stable} as our base model. Similar to~\cite{Mei2024CVPR,mei2025lux}, we decompose the relighting task into two steps: first delight the subject to flat-lit albedo, and then relight the albedo to the target lighting. More specifically, given a portrait under arbitrary lighting condition, we first apply a delight model to cancel out the portrait's original lighting effect and estimate its flat-lit albedo. We then apply a relight model that takes in the albedo video, and uses environment map and background image as lighting control to generate a relit video under the target lighting condition. 

We adopt the standard training pipeline to fine-tune SVD with a diffusion U-Net. The input video $\mathbf{V}_i$ is first encoded into a latent representation $z = \mathcal{E}(\mathbf{V}_i)$ by a Variational Auto Encoder (VAE). It is then concatenated with a noise latent $z_t$ (where $t$ is the time step) to feed into a diffusion U-Net\footnote{Here $\mathbf{c}$ is an optional embedding for conditioning the U-Net. We set $\mathbf{c}$ as a dummy condition.}: $\mathrm{f}_\theta (z_t; \mathbf{c}, t)$ that is able to iteratively remove noise from the noise latent to recover a clean latent $z_o$, which can be further decoded into the desired output video. The diffusion U-Net is trained by minimizing the denoised latent and ground truth latent at each time step: 
\begin{equation}
\min_\theta||\mathrm{f}_\theta (z_t; \mathbf{c}, t) - z_o ||.
\end{equation}

Both our delight and relight models are trained with this similar pipeline. The relight model further incorporates an environment map and a background image to control the lighting of the generated relit video.

\paragraph{Delight Model} The goal of the delight model is to estimate the flat-lit albedo under a pure white environment light, given an arbitrarily lit portrait video. This step simplifies the relight problem and makes it more tractable.

The delight model is trained with portrait videos $\mathbf{V}$ taken under different environment maps and their paired albedo reference videos $\mathbf{A}$. To train the model, $\mathbf{V}$ is concatenated with noise latent and used as input to the diffusion U-Net. $\mathbf{A}$'s latent encoding: $z_o^A = \mathcal{E}(\mathbf{A})$ is used as the ground truth latent. In the inference phase, the delight model takes an arbitrarily lit portrait video and synthesizes its flat-lit albedo video. Since we use video diffusion U-Net with temporal convolution and attention layers, the output video exhibits good temporal consistency within the trained sequence length. 

\paragraph{Relight Model} The relight model takes a flat-lit albedo video $\mathbf{A}$ and estimates the relight video $\mathbf{V}_r$, given a desired environment light. It is trained using the same diffusion U-Net backbone. The environment map is injected as an explicit lighting condition following the design of DiffusionRenderer~\cite{DiffusionRenderer}. Since the environment maps typically are not pixel-aligned with the input, direct concatenation with the input latent~\cite{jin2024neural_gaffer} would result in sub-optimal control effect. Here we use encoded environment map as lighting control through cross-attention layers. Specifically, the environment maps are first encoded by a VAE encoder into a latent vector: $z^E = \mathcal{E}(\mathbf{E})$. We then use another environment map encoder to further process $z^E$ in order to extract a set of multi-resolution feature maps: $\mathbf{c} = \mathcal{E}_{env}(z^E)$. This encoder follows a simplified diffusion U-Net encoder architecture, consisting of several convolutional layers for progressive downsampling, but with the attention and temporal modules removed. Those multi-resolution features are injected into the multi-level cross-attention layers as lighting control.

To train the relight model, we composite the albedo video with the desired background image: $ M \cdot A + (1-M) \cdot B$ (where $M$ is the foreground matting mask, $B$ is the background image, and this equation indicates per-frame operation), and concatenate it with the noise latent. The matting operation is more efficient than directly concatenating the background image by using fewer input channels. The environment map $\mathbf{E}$ first goes through a VAE encoder and then an environment map encoder. The encoded environment map latent is then passed into the multi-level cross-attention layers. We use the latent encoding of video taken under $\mathbf{E}$ as the ground truth latent: $z_o^E = \mathcal{E}(\mathbf{V}_E)$ (where $\mathbf{V}_E$ is the video whose lighting is consistent with $\mathbf{E}$). In the inference stage, the relight model takes a background-composited albedo video and an environment map and outputs a relit portrait video with lighting consistent with the environment. Note that when a background image is not available, we can use camera parameters to calculate a projected background from the environment.

\paragraph{Lighting Conditioning} In the relight model, we use an environment map and a background image to control the lighting effect in the relit output. Since the environment is full-surround, the spatial relationship between the subject and environment needs to be specified. We use the camera extrinsic parameters to rotate the environment and align it with the camera's look-at direction, such that the camera always looks at the center of the environment after the rotation. Given the camera's intrinsic parameters, we can project the environment to a background image. To accommodate the data range of an HDR environment map, we convert it into three images: one tone-mapped LDR image to preserve the color of the lighting, one normalized log-intensity image to preserve the HDR contrast, and one directional encoding image to represent the lighting direction in the camera coordinate system. The three images are concatenated to feed into the VAE encoder. 

The background image is combined with the albedo input to provide lighting conditioning. The background image can control the overall color tone and brightness level of the relit portrait such that it looks more harmonious with the background.

\paragraph{Long Video Inference} Since the video diffusion model is trained with a fixed length, inferencing long videos that exceed the length would suffer from inconsistency as the long video needs to be divided into multiple clips to fit the fix length. In order to relight long video while maintaining overall temporal consistency, we adopt a sliding window scheme that divides the long sequence into overlapped short sequences. For the overlapped frames, we take the noise from the previous sequence to ensure the temporally consistent generation of consecutive frames in the current sequence. Specifically, given a noise latent $z_t = z_0 + \sigma_t\epsilon$ (where $z_0$ is the clean latent, $\epsilon$ is a Gaussian noise, and $\sigma_t$ is a per time step coefficient that follows the EDM noise schedule~\cite{Karras2022edm}), we take the $z_0$ and $\epsilon$ from the previous sequence for the overlapped frame of the current sequence. This ensures that the overlapped frames are identical in the two neighboring batches. By leveraging the temporal attention within the fixed length sequence, we can achieve temporally consistent inference of long videos. In our experiments, we train a 30-frame diffusion model and overlap 20 frames when inferencing long videos. 

\begin{table*}[tb]
  \caption{\textbf{Quantitative comparisons on the delight and relight results.}}
  \label{tab:quan_comp}
  \vspace{-6pt}
  \centering
  \begin{tabular}{c|c| c c c| c c c| c c c| c c c}
    \Xhline{1pt}
    \multirow{2}{*}{Task} &
    \multirow{2}{*}{Method} &
    \multicolumn{3}{c|}{Subject 1} &
    \multicolumn{3}{c|}{Subject 2} &
    \multicolumn{3}{c|}{Subject 3} &
    \multicolumn{3}{c}{Subject 4} \\ \cline{3-14}
     & & PSNR$\uparrow$ & SSIM$\uparrow$ & LPIPS$\downarrow$
    & PSNR$\uparrow$ & SSIM$\uparrow$ & LPIPS$\downarrow$
    & PSNR$\uparrow$ & SSIM$\uparrow$ & LPIPS$\downarrow$
    & PSNR$\uparrow$ & SSIM$\uparrow$ & LPIPS$\downarrow$ \\
    \Xhline{1pt}

    \multirow{4}{*}{Delight}
    & PN-Relight   
        & 17.55 & 0.90 &  0.09
        & 16.56 & 0.89 & 0.10
        & 16.76 & 0.89 & 0.10 
        & 19.77 & 0.93 & 0.06 \\
    \cline{2-14}
    & SwitchLight  
        & 20.81 & 0.90 & 0.12 
        & 18.48 & 0.88 & 0.14
        & 19.40 & 0.91 & 0.07
        & 21.34 & 0.93 & 0.05 \\
    \cline{2-14}
    & RelightVid   
        & 7.18 & 0.78 & 0.28 
        & 6.54 & 0.70 & 0.32
        & 8.00 & 0.74 & 0.23 
        & 7.12 & 0.70 & 0.29 \\
    \cline{2-14}
    & \textbf{Ours}
       & \textbf{25.20} & \textbf{0.93} & \textbf{0.07}
      & \textbf{24.62} & \textbf{0.93} & \textbf{0.10}
      & \textbf{29.73} & \textbf{0.93} & \textbf{0.06}
      & \textbf{28.25} & \textbf{0.94} & \textbf{0.05}\\
   \Xhline{1pt}

    \multirow{4}{*}{Relight}
    & PN-Relight   & 14.33 & 0.77 & 0.27 & 13.80 & 0.71 & 0.28 & 14.14 & 0.72 & 0.24 & 14.43 & 0.75 & 0.25 \\
    \cline{2-14}
    & SwitchLight  & 15.93 & 0.68 & 0.28 & 14.82 & 0.78 & 0.21 & 12.88 & 0.68 & 0.22 & 13.88 & 0.68 & 0.18 \\
    \cline{2-14}
    & RelightVid   & 10.08 & 0.70 & 0.33 & 20.50 & 0.83 & 0.27 & 17.60 & 0.68 & 0.25 & 11.12 & 0.62 & 0.26 \\
    \cline{2-14}
    & \textbf{Ours}
      & \textbf{21.50} & \textbf{0.87} & \textbf{0.18}
      & \textbf{20.84} & \textbf{0.88} & \textbf{0.17}
      & \textbf{23.05} & \textbf{0.87} & \textbf{0.14}
      & \textbf{23.75} & \textbf{0.88} & \textbf{0.14} \\
  \Xhline{1pt}
  \end{tabular}
\end{table*}

\section{Evaluation and Results}
We evaluate the performance of our approach on real-world portrait videos. We first test on the videos captured in our own system for both quantitative and qualitative evaluation. We then test on the in-the-wild online videos to demonstrate the generalizability of our model. We compare our results with several state-of-the-art video relighting methods, whose codes or demos are available online.

\paragraph{Implementation Details}
We train our models using our hybrid dataset, with ground-truth flat-lit albedos, environment maps, and background images. We exclude the data of four real-captured subjects (such that they are not seen by our models during training), so we can use them for quantitative evaluation. Due to the scarcity of video relight data with ground-truth, we will make the evaluation data publicly available for benchmarking video relighting algorithms. We train two models for delight and relight. Both models are initialized using the pre-trained SVD~\cite{blattmann2023stable}. The two models are trained separately with the same setting. We train the models with two lengths: 10 frames and 30 frames. We use a GPU node with four A100 and 100G memory for training. The batch size we use for the 10-frame model is 3 per GPU with gradient accumulation step 8. For the 30-frame model, we reduce the batch size to 1 per GPU due to the limit of GPU memory. In both modes, the models are trained for 10K iterations with learning rate $1e^{-4}$. We use the EDM noise scheduler with 1,000 time steps for training. The training time for both modes is around 16 hours. In the inference stage, we denoise each frame for 50 steps. It takes around 1 seconds to inference 1 video frame once the model is loaded.

\paragraph{Evaluation on Our Captured Data} 
We first test on the data captured in our own stage. Since we capture time-multiplexed motion sequences under two known environment maps, we can use our data for pixel-aligned quantitative evaluation. Specifically, we use one sequence as the input source video and the other as ground-truth relit video for its target environment. We use four subjects, whose images are not included in the training data. For each subject, we relit with four different environment maps. We first delight the input video, and then relight its albedo using the target environment map. Since the sequence we use for this experiment is short, we use the 10-frame model for inference. Although we also have synthetic data paired with ground-truth lighting and albedo, we choose not to use synthetic data for evaluation due to the sim-to-real gap. And therefore the evaluation on synthetic data may not reflect the model's actual performance. 

\begin{figure*}[t]
  \includegraphics[width=1\linewidth]{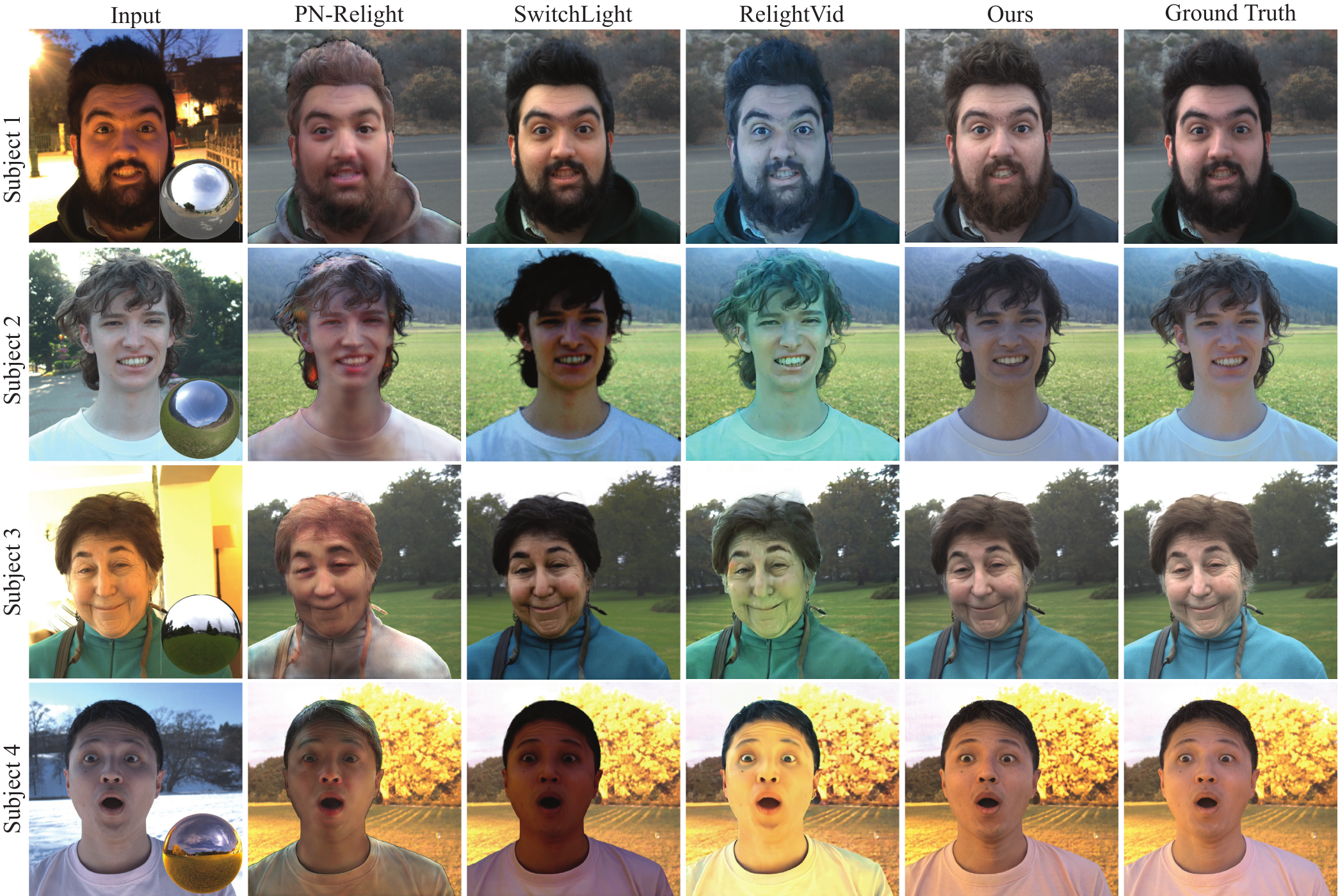}
  \caption{\textbf{Relight visual comparison results.} For each subject, we show an input frame with target environment. We show our relit results in comparison with PN-Relight~\cite{wang2023free}, SwitchLight~\cite{kim2024switchlight}, RelightVid~\cite{fang2025relightvid}, and the ground-truth.}
  \label{fig:real_result_1}
\end{figure*}

We compare our delight and relight results with three state-of-the-art relighting methods: PN-Relight~\cite{wang2023free}, SwitchLight~\cite{kim2024switchlight}, and RelightVid~\cite{fang2025relightvid}. Both PN-Relight and SwithLight are CNN-based methods. PN-Relight is trained with a small OLAT dataset for single-image free-viewpoint relighting. Their model is an image-based model and does not incorporate temporal information. We run experiments with their inference code and pre-trained model provided on their website. SwitchLight (or Beeble\footnote{https://beeble.ai}) is a commercial AI tool for portrait relighting. We use their web interface to generate the delight and relight results by uploading our data. RelightVid is a recent diffusion-based method for general object relighting. We use its publicized code and pre-trained model to run our experiments. We compare with these methods on both delight and relight results. For SwitchLight, we use its generated ``base color" as the albedo result. For PN-Relight and RelightVid, since they don't directly provide a delight model, we use pure white environment as input to these methods to generate their delight results. 

\begin{figure*}[t]
  \includegraphics[width=1\linewidth]{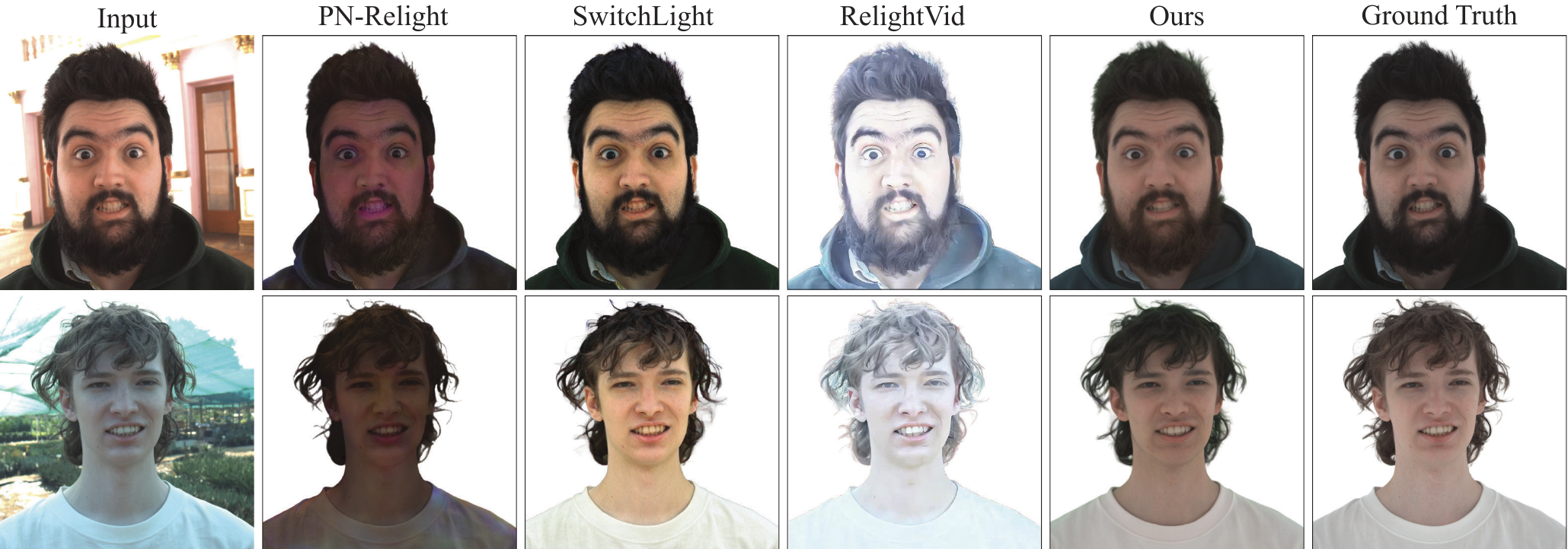}
  \caption{\textbf{Delight visual comparison results.} We show our relit results in comparison with state-of-the-art relighting methods: PN-Relight~\cite{wang2023free}, SwitchLight~\cite{kim2024switchlight}, and RelightVid~\cite{fang2025relightvid}. }
  \label{fig:delight_result}
  \vspace{-8pt}
\end{figure*}

 \begin{figure*}[t]
  \includegraphics[width=1\linewidth]{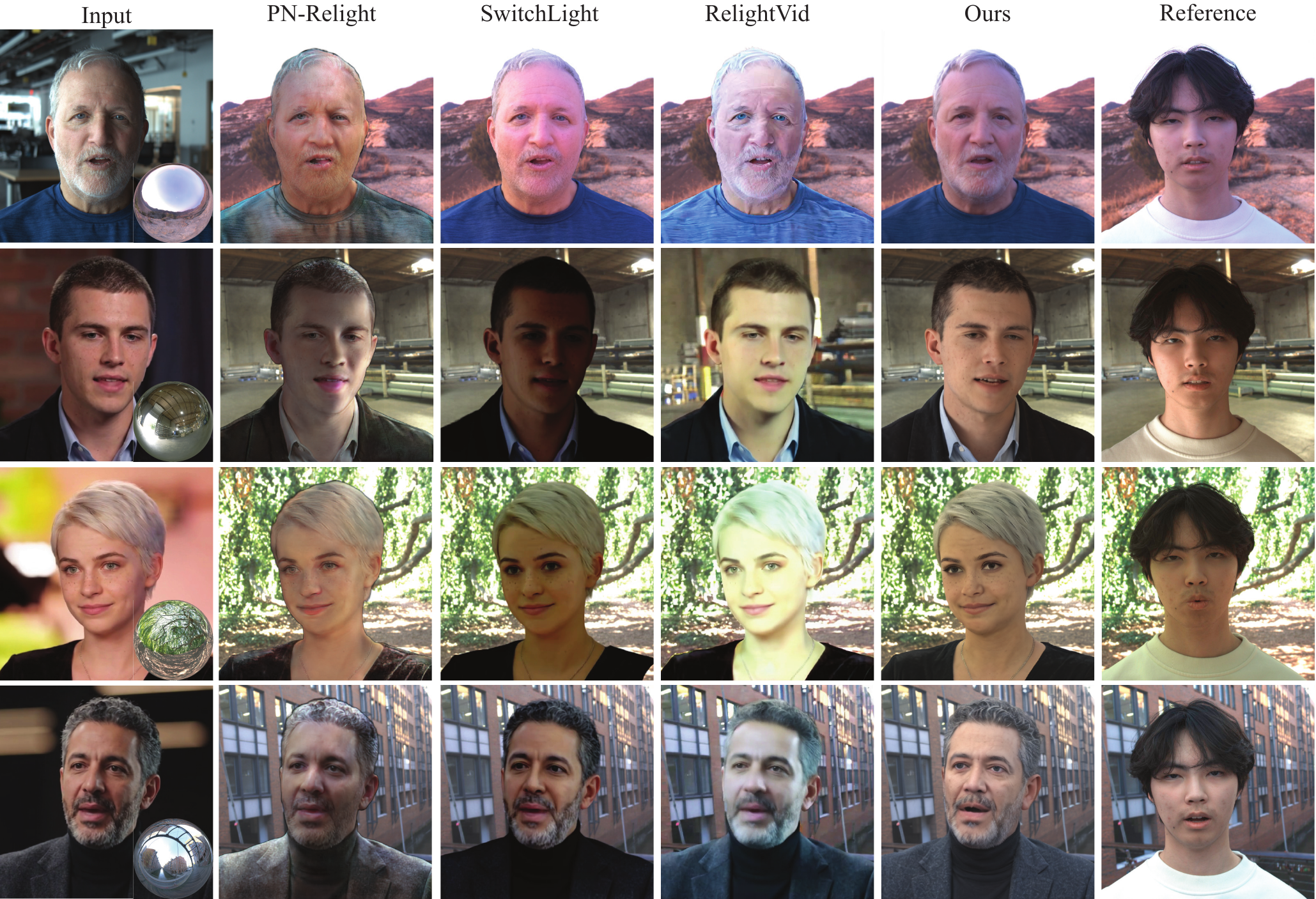}
  \caption{\textbf{Visual comparison on in-the-wild videos.} We show our relit results on in-the-wild videos in comparison with state-of-the-art relighting methods: PN-Relight~\cite{wang2023free}, SwitchLight~\cite{kim2024switchlight}, and RelightVid~\cite{fang2025relightvid}. The first row shows an input taken by ourselves, and the other three rows show inputs from online videos. We show one input frame and the target environment map in the first column. Since we do not have ground-truth relit results for in-the-wild inputs, we show reference images of a subject taken in the Pixel Cube using the same environment map.  }
  \label{fig:relight_wild}
\end{figure*}

The visual comparisons for the delight and relight results are shown in Fig.~\ref{fig:delight_result} and ~\ref{fig:real_result_1} respectively. Video results are available in the supplementary video. We show the input image, delight or relight results of each method, and the ground-truth images captured in the Pixel Cube. For the relighting results, we also show the target environment map. We can see that our results resemble the ground-truths the most in both the delight and relight tasks. As a commercial software, SwitchLight generates impressive results that also well preserved the subject's identity. However, their lighting consistency is lacking in some cases (for example, the relight results of subject 2) and their matting masks are sometimes inaccurate when the boundary is fuzzy (see subject 2's hair region). The RelightVid results look unrealistic as their model is trained with general objects and is not tailored for portrait relighting.

We perform quantitative evaluation on the results by comparing with our ground-truth data. Specifically, we use three reference-based metrics, PSNR, SSIM, and LPIPS, for evaluation. Both PSNR and SSIM calculate per-pixel similarity between two images. LPIPS is a perceptual-based metric that evaluate how similar two images look to humans. All metrics are calculated on the portrait region only, since the relit results are composite to the same background image using matting masks. The metric comparison for both the delight and relight results are reported in Table~\ref{tab:quan_comp}. For each subject, the relight metrics are averaged over four different environment map. We can see that our models achieve the best performance for all subjects on both tasks, which is consistent with the visual comparison results. Although SwitchLight's visual results well preserve the subjects' appearance, their PSNR is relatively low since many of the results are much darker than the ground-truth, resulting low similarity scores. The PSNR of RelightVid relight result is higher than SwitchLight, although they are visually lacking, since the RelightVid results have color balancing issue: their blue channel values are too high, while the values of the other two channels are closer to the ground-truth than SwitchLight ones. Since the overall color tone of SwitchLight results is better than RelightVid, they have better performance on perceptual metric (\ie, LPIPS), although the PSNR score lower.

\paragraph{Evaluation on In-the-wild Data.} We then perform experiments on portrait videos taken in the in-the-wild environment. We record such videos by ourselves and also download from online. In these experiments, we test on longer videos and use our 30-frame model for inferencing. For videos longer than 30 frames, we use our long sequence inference scheme to delight/relight the entire video. We test on various subjects with different genders, ages, and skin tones. Since the ground-truth relit data is not available, we cannot perform quantitative evaluation and only demonstrate visual results. For those target environment maps used in our system, we show an example of our captured subject as reference to the expected lighting effect. We perform comparison with the three state-of-the-art relight models used in the previous experiment. The visual comparisons of the in-the-wild video relight results are shown in Fig.~\ref{fig:relight_wild}. We can see that our model generalizes on in-the-wild videos very well, thanks to the Pixel Cube for reproducing realistic lighting for training data acquisition. Our results have similar lighting effect to the reference images, in terms of shadow and highlight distributions. SwitchLight results are sometimes inconsistent with the reference. The subject's identity, even their wrinkles and facial hair, are preserved very well in our results, whereas some other methods distorts the subjects' appearance. 

\begin{figure*}[t]
  \includegraphics[width=0.99\linewidth]{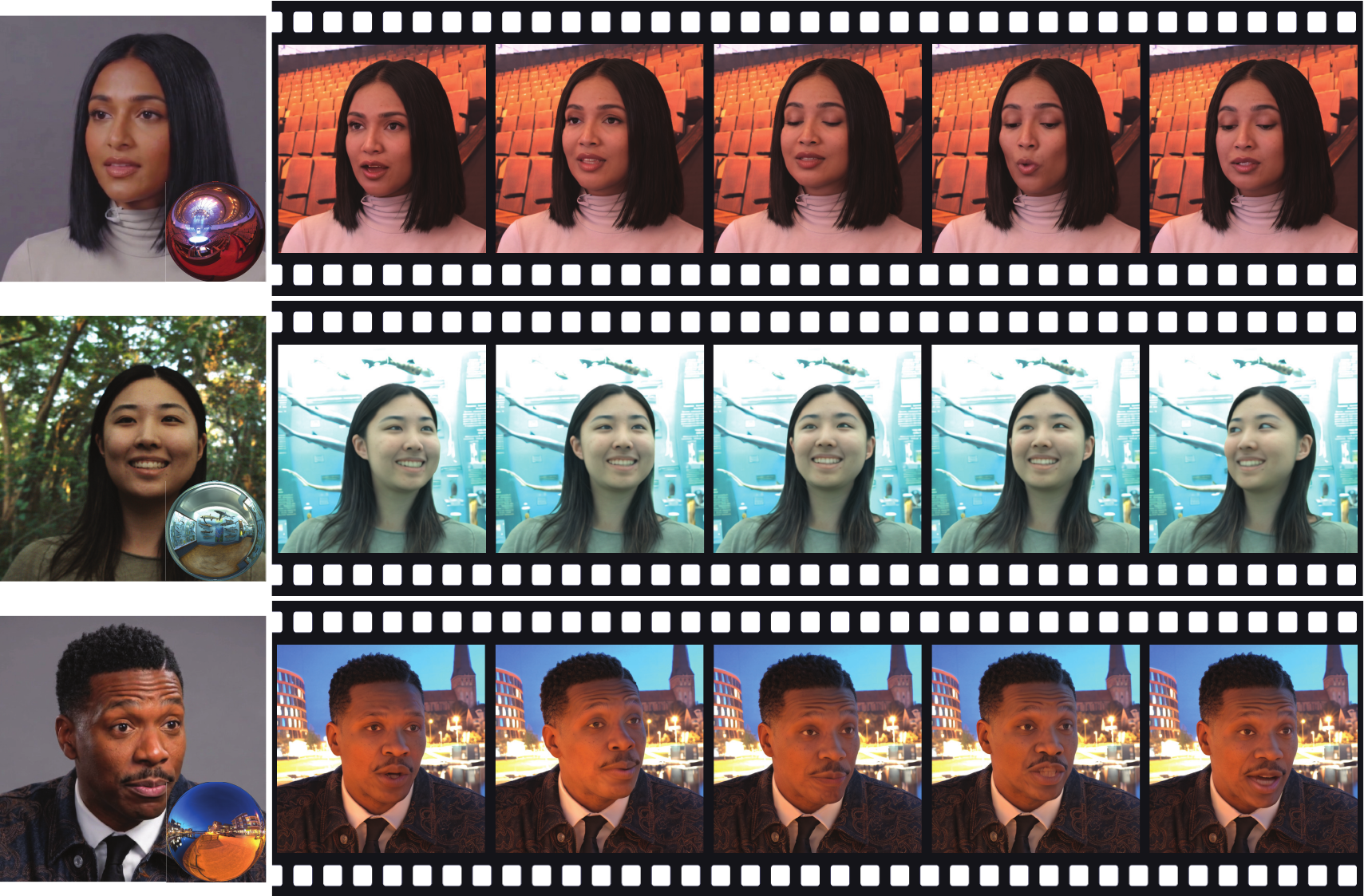}
  \vspace{-6pt}
  \caption{\textbf{Dynamic relit results on in-the-wild videos.} We show one input frame with the target environment and five frames from our relit video. }
  \label{fig:relight_seq}
\end{figure*}

Fig.~\ref{fig:relight_seq} shows our relight result with multiple temporal frames. We can see that our result has good temporal consistency that is free from flickering and can be adapted to large head motions (for example, the girl in the first row sways her head from right to left). This is achieved by using multi-view images for training. 

\begin{figure*}[t]
  \includegraphics[width=0.99\linewidth]{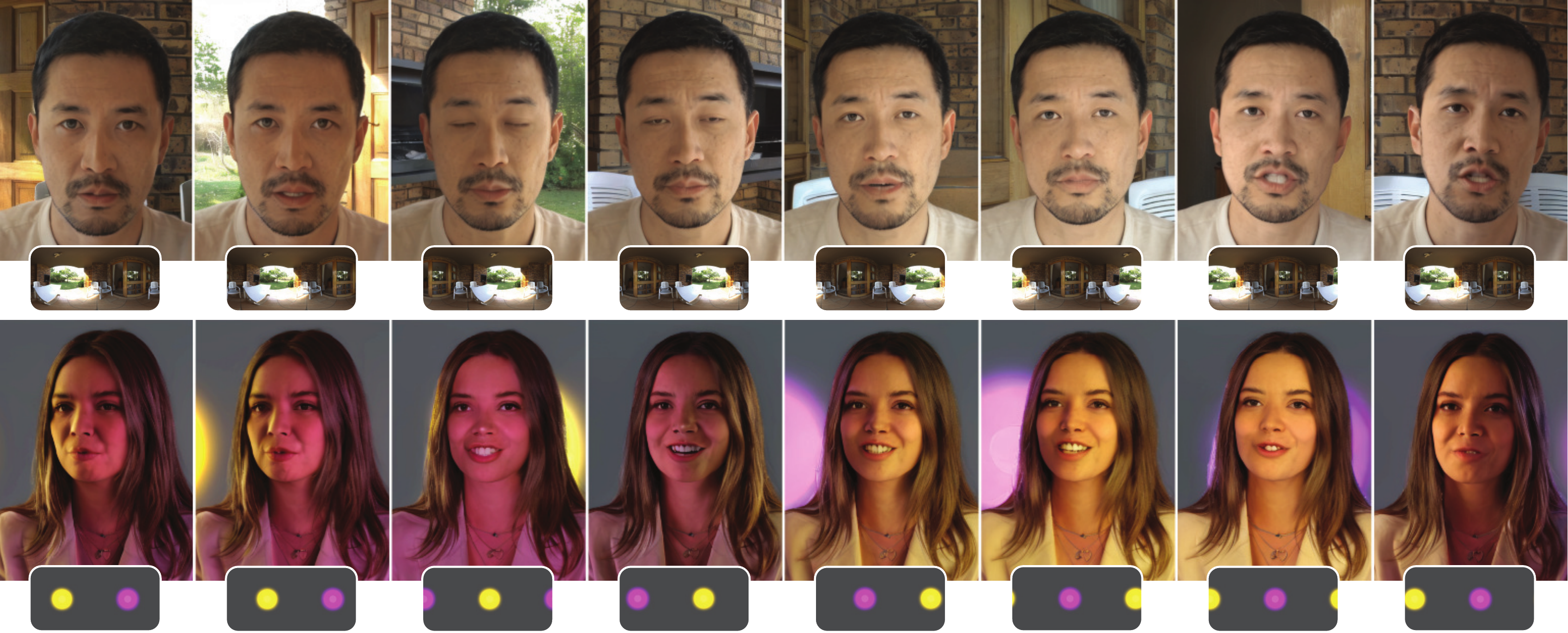}
  \vspace{-6pt}
  \caption{\textbf{Dynamic video relighting results under rotating environment light.} We show example frames from two in-the-wild relit videos with horizontally rotating environment lighting.}
  \label{fig:rot}
\end{figure*}

Fig.~\ref{fig:rot} shows two dynamic relighting results with rotating environment light. In this experiment, both the subject and the environment map are in motion. Here we rotate the environment for $360^\circ$ with a $12^\circ$ interval each step. The sequence has 30 frames in total. We can see that our model can still produce lighting consistency and temporal coherence results in this challenging experiment. The example in the second row has highly contrastive lighting, our model successfully reproduce the high contrast appearance with consistent color to the environment light. The dynamic video results of the above examples are available in our supplementary video.

 \begin{figure*}[t]
  \includegraphics[width=\linewidth]{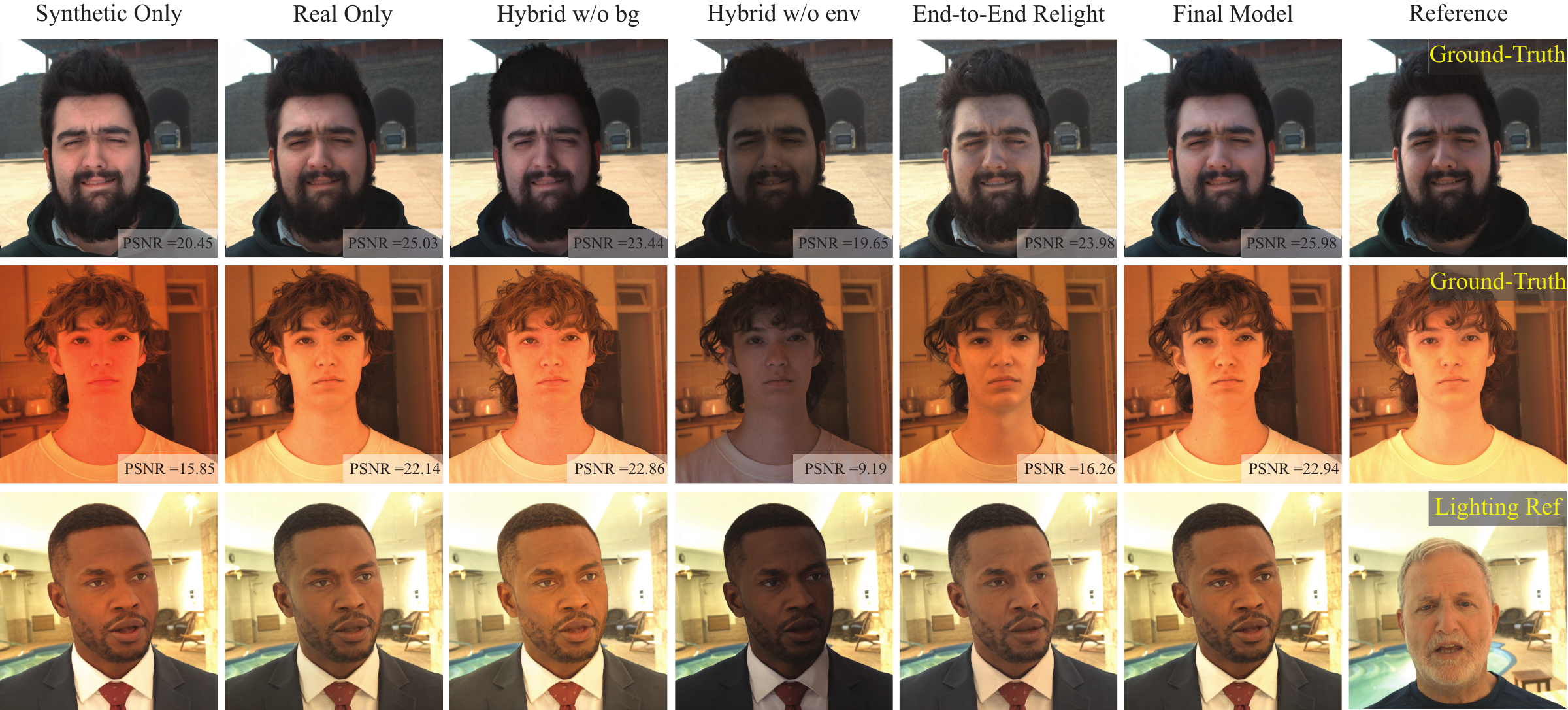}
  \caption{\textbf{Ablation results.} Here we show relighting results using different variants of our model. The first two rows are from our real-captured data with ground-truth relit reference. The third row is in-the-wild data. We show a reference image taken under the same environment lighting.}
  \label{fig:ablation}
\end{figure*}

\begin{figure}[t]
  \includegraphics[width=\linewidth]{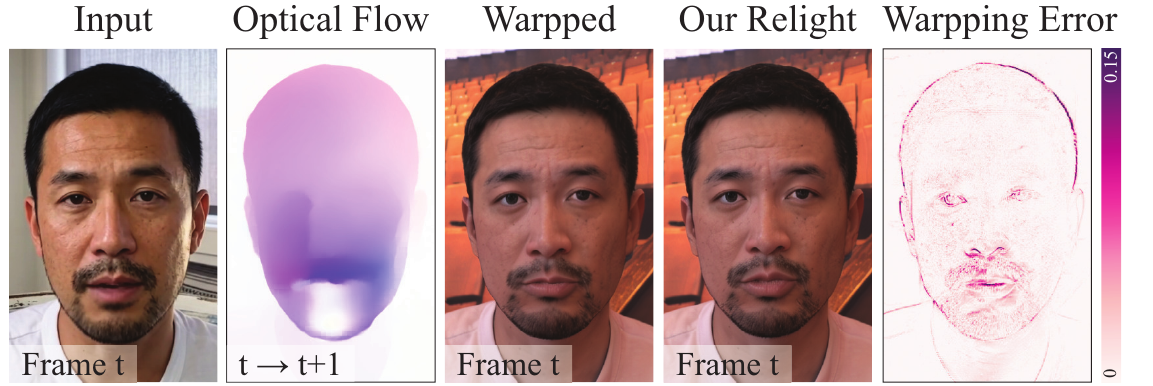}
    \vspace{-8pt}
  \caption{\textbf{Temporal warping error.} We evaluate the difference between our inferred relit frame and a warped frame using the optical flow estimated from the input.}
  \label{fig:warping}
\end{figure}

\paragraph{Ablation Studies.} We perform ablation studies to evaluate the design choices of our model. Specifically, we evaluate the choice of training data type, lighting control, and our overall delight-then-relight pipeline. We compare our final model with five variants trained in different ways: trained on synthetic data only (``Synthetic Only''), trained on real-captured data only (``Real Only''), trained on hybrid data without using the environment map control (``Hybrid w/o env''), trained on hybrid data without using the background control (``Hybrid w/o bg''), and an end-to-end model that directly relight without the delight step (``End-to-End Relight''). Our final model is trained on the hybrid data using both the background image and environment map as lighting control with the delight-then-relight pipeline. Visual comparison results are shown in Fig.~\ref{fig:ablation}. We show two examples from our own data and one example from in-the-wild data. Since our own data has ground-truth relit images for quantitative evaluation, we show the ground-truth as reference and also the PSNR score of each relit result. For the in-the-wild data, we provide a reference portrait image captured in the Pixel Cube under the same environment light. We can see that although the results of all variants look visually pleasing, our final model achieves the highest PSNR, indicating the closest resemblance to the ground-truth. The model trained with synthetic data falls short in relighting real-captured videos. The environment map imposes strong lighting control via multi-level cross-attention. We can see that without using the environment map, the model cannot effectively control the lighting and the relit images appear dark. By concatenating the background image, the overall color tone is further adjusted and becomes more consistent with the environment. The end-to-end trained model's performance tends to be sensitive to input brightness. For example, its relit result for subject 1 appears to be brighter than the ground-truth, whereas the subject 2's relit result is darker. This is due to the input's lighting variation. In contrast, our delight-then-relight pipeline first normalize the subject's appearance by delighting and thus achieves more stable performance in relighting.

 \begin{figure}[t]
  \includegraphics[width=\linewidth]{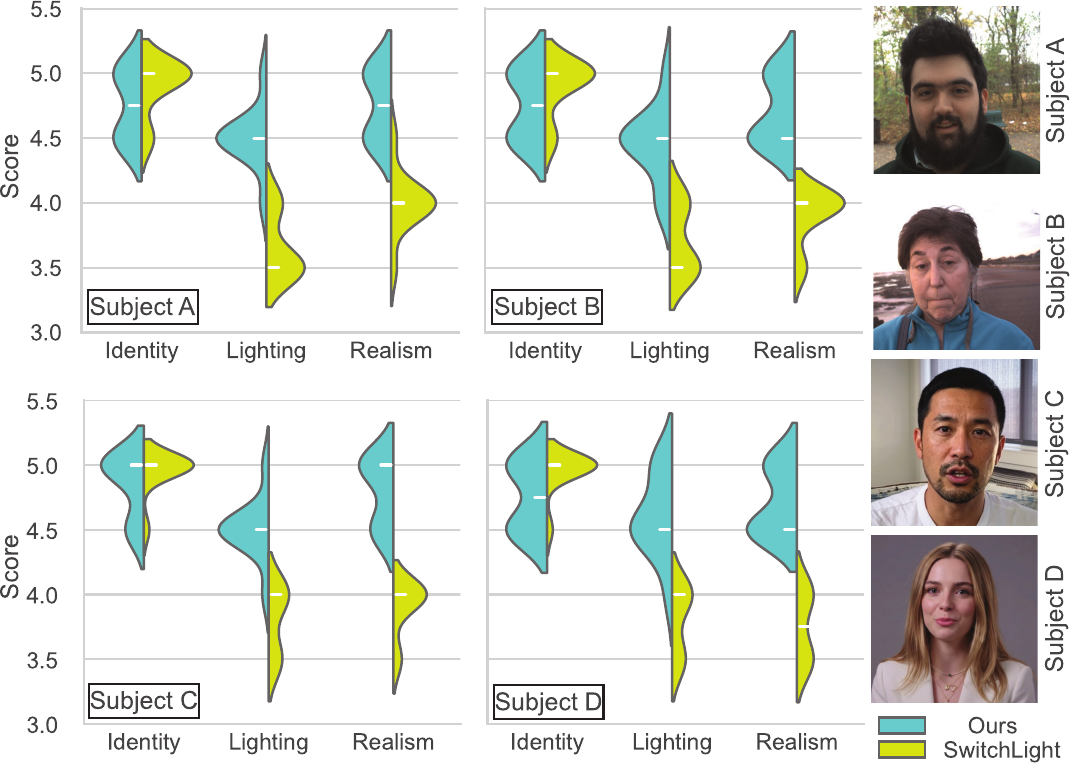}
  \caption{\textbf{User study results.} Here we show the density distribution of user ratings with respect to identity, lighting consistency, and realism. The white tick indicates the median. The four subjects used as input are shown on the right.}
  \label{fig:user_study}
\end{figure}

\begin{figure*}[t]
  \includegraphics[width=\linewidth]{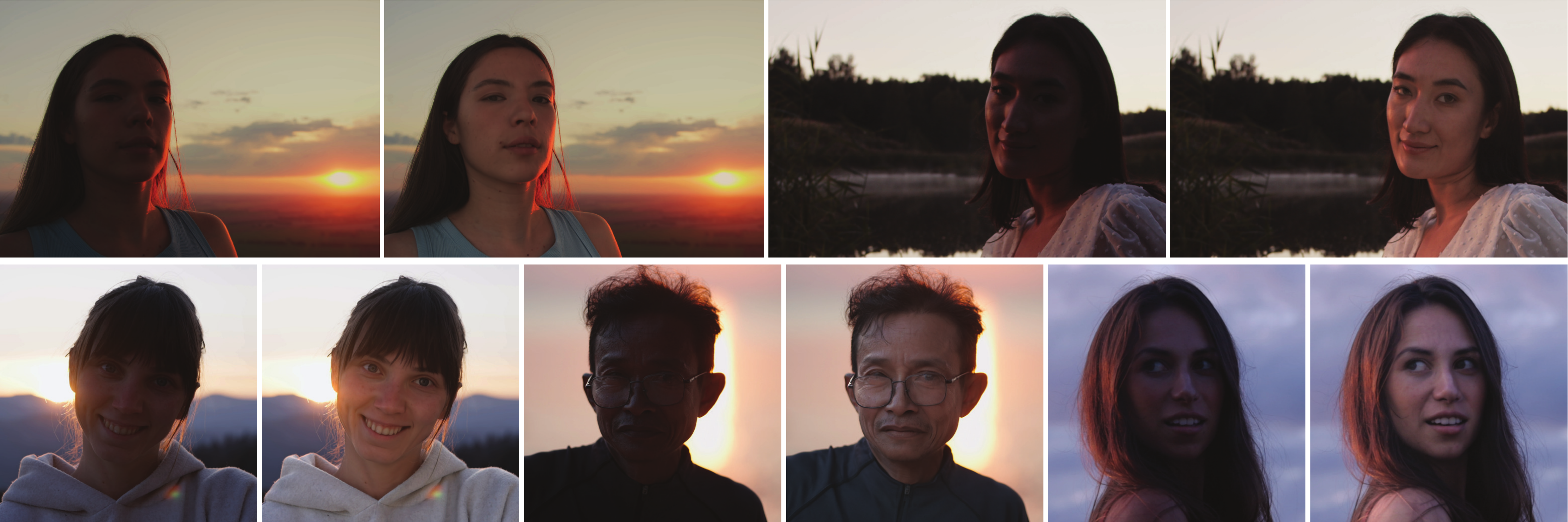}
    \vspace{-18pt}
  \caption{\textbf{Under-exposed portrait enhancement.} Here we use our relighting model to improve the illumination in under-exposed portrait photos. Third-party material sources: SBV-349294699, SBV-348642689, SBV-352471750, SBV-353602821, and SBV-352073197 from \url{Storyblocks.com} [Licensed under author's individual subscription].}
  \label{fig:fill_light}
    \vspace{-6pt}
\end{figure*} 

\begin{figure}[t]
  \includegraphics[width=\linewidth]{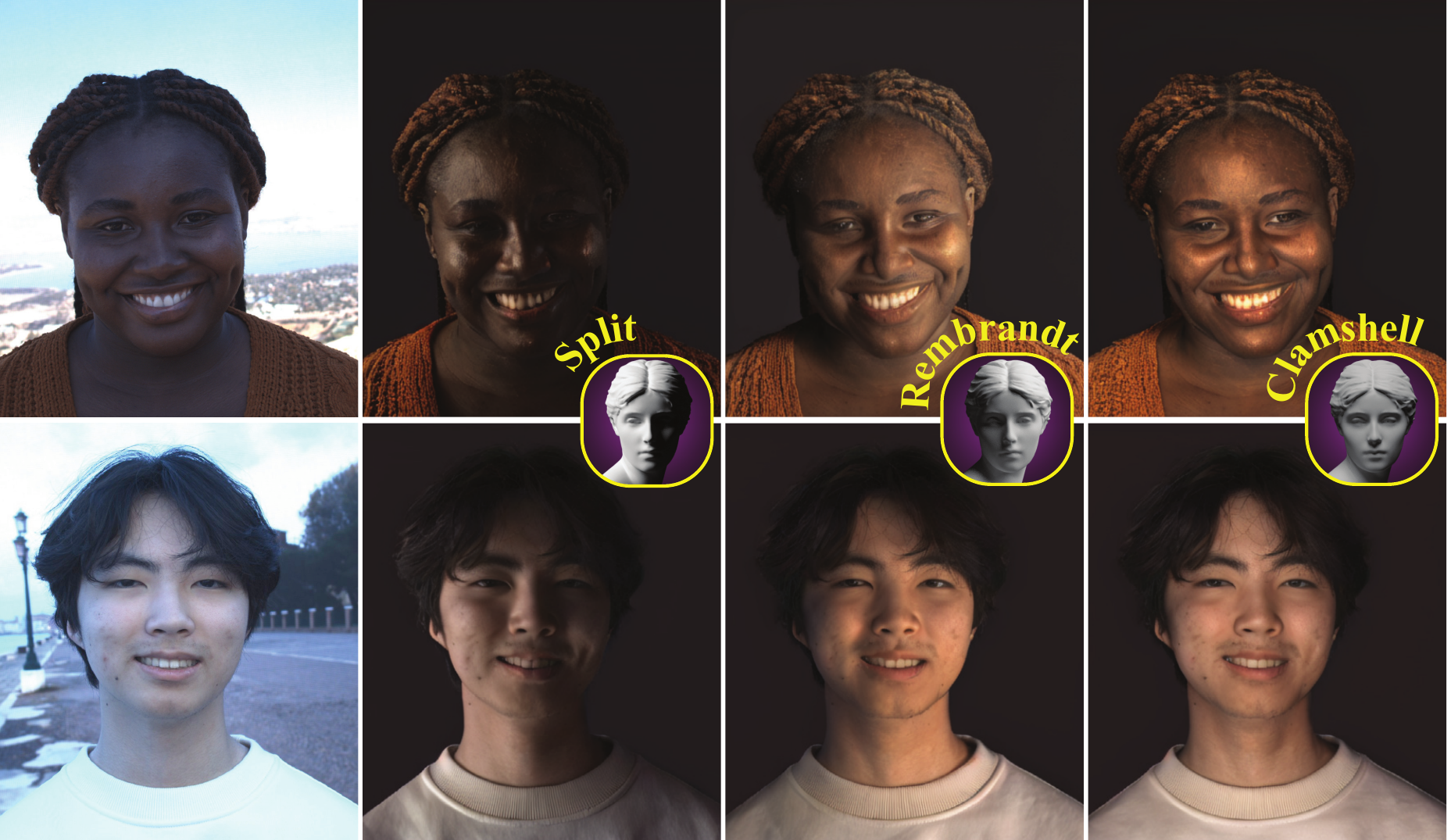}
    \vspace{-18pt}
  \caption{\textbf{Professional portrait lighting.} Here we show portrait images relit using our model under three professional lighting setups: split light, Rembrandt, and Clamshell. The input images are shown in the first column.}
  \label{fig:effect}
    \vspace{-6pt}
\end{figure}

\paragraph{Evaluation on temporal consistency.} To quantitatively evaluate the temporal consistency of our relight result, we adopt the temporal warping error based on the input optical flow. Specifically, we estimate the dense optical flow using two neighboring frames of the input video. We then use the optical flow to warp our relight frame to obtain its neighboring frame. We compute the error between the warped frame and our relit frame at the same time instance. The warping error of one sample frame is shown in Fig.~\ref{fig:warping}. We can see that the errors are under 0.12. We perform this evaluation on ten sequences of different subjects. The average temporal warping error across all sequences is 0.005. This indicates that our relighting result exhibits the same motion as the input sequence, with reliable temporal consistency.

\paragraph{User Study}
We conduct a user study to evaluate the perceptual quality of our relighting results in comparison with the SwitchLight results. Specifically, we recruit ten participants with ages ranging from 21 to 45. We show them the input video and relit video side-by-side and ask them to rate the relit result with a 5-point scale, where 5 indicates the most positive, 3 is neutral, and 1 the most negative.

We use videos of four subjects for this experiment (two from our own data and two from in-the-wild, see Fig.~\ref{fig:user_study}). Each video is relit using 5 different environment maps. Therefore each participant is presented with 20 videos. For each relit video, we ask the participant to provide ratings based on the following three questions: 1) Identity: does the person that appears in the two videos seem to have the same identity? 2) Lighting: in video 2 (\ie, the relit video), does the person's facial lighting appear to be consistent with the background environment? and 3) Realism: does video 2 appear to be a real captured video clip? The user ratings are summarized in Fig.~\ref{fig:user_study}. We show the density distribution of the ratings along with the median. We can see that SwitchLight results achieve high scores on preserving the identity, since they relight using the target's geometry and albedo without a generative process. Although being slightly lower, our identity ratings are all positive and comparable to SwitchLight. In terms of lighting consistency and overall realism, our relit results achieve significantly higher ratings than SwitchLight.

\section{Applications} 
Lastly, we demonstrate using our relighting model for practical photographic and cinematographic use cases. Specifically, we showcase three applications: under-exposed portrait enhancement, professional portrait lighting, and video lighting unification. Please see the supplementary video for dynamic results of these applications. 

\begin{figure}[t]
  \includegraphics[width=\linewidth]{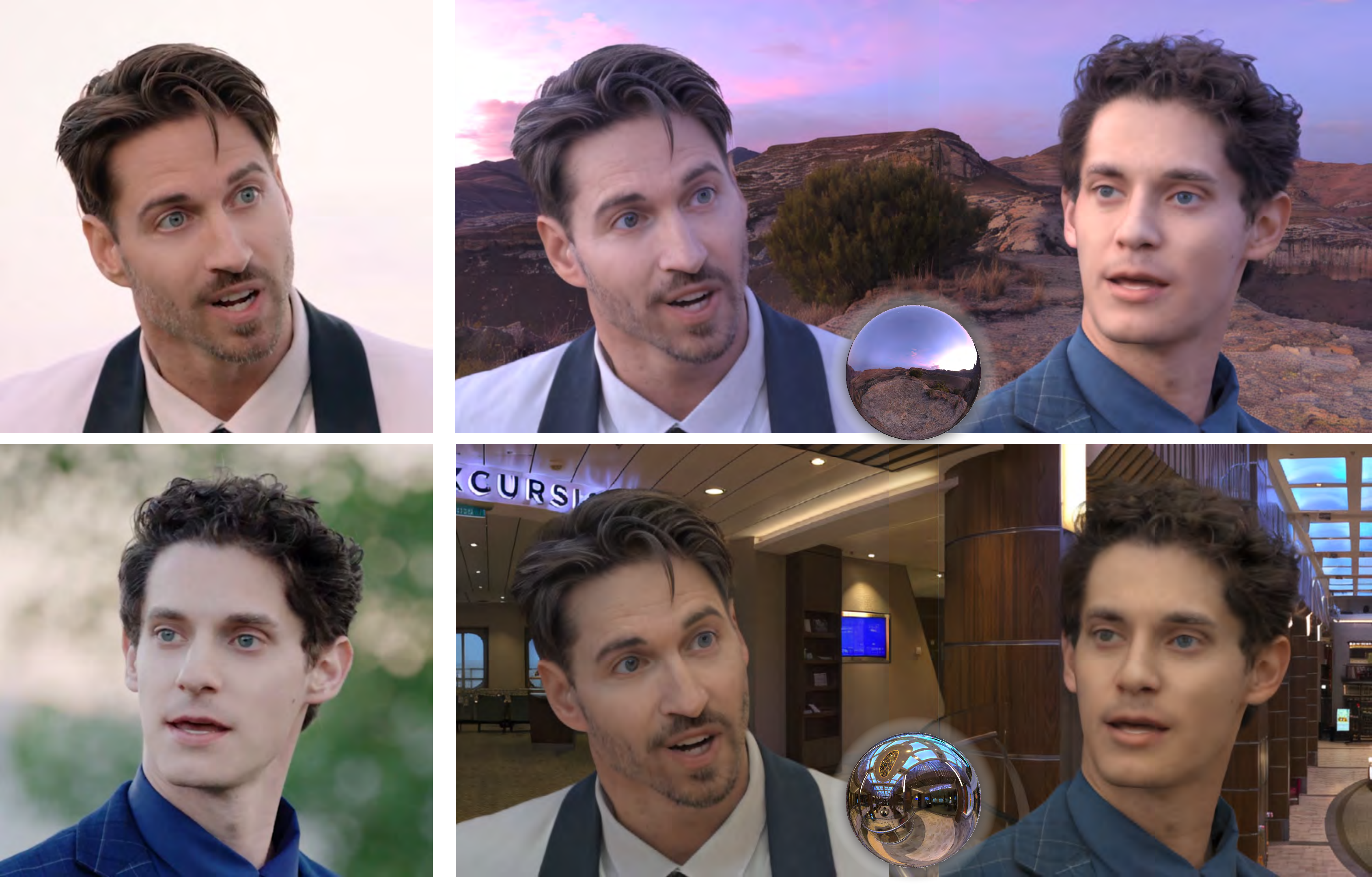}
    \vspace{-24pt}
  \caption{\textbf{Video lighting unification.} Here we show two actors filmed under different lighting conditions (left). They are relit to a unified illumination environment (right) using our model. Source videos courtesy of Journey Entertainment.}
  \label{fig:short_video}
\end{figure}

\paragraph{Under-exposed Portrait Enhancement.} Under-exposed portrait images are usually undesirable, as the facial features may appear obscured. We can use our model to relight the subject to enhance the facial lighting, without changing the original background. This is equivalent to lighting up the subject with additional fill lighting that is commonly used in portrait photography. We achieve this by relighting the subject with an environment map that has a point light source facing the subject and then blend the relit image with the original photo. In this way, we enhance the face lighting without changing the background. Fig.~\ref{fig:fill_light} shows our portrait enhancement results.

\paragraph{Professional Portrait Lighting.} Professional portrait photography usually requires a complex setup of lighting fixtures in order to achieve dramatic lighting effect. Such setup can be daunting to average people without professional photographic skills. Our model can make professional portrait photography accessible to everyone by allowing post-relighting without requiring physical lighting equipment during shooting. One can designate the environment map for a desired lighting effect and relight an arbitrary portrait photo using our model. Example results of professional portrait relighting are shown in Fig.~\ref{fig:effect}. Here we show three lighting patterns with key light from different directions. We can see that our results achieve the desirable lighting effects and well preserve the target's identity.

\paragraph{Video Lighting Unification.} Outdoor filming often faces the issue of inconsistent lighting as the sunlight changes over time. This is problematic when shooting a scene that involves multiple actors, but each of them is lit under a different lighting condition. This problem can be resolved by using our model to relight the actors under the same environment light, such that their lighting effects are unified. Fig.~\ref{fig:short_video} demonstrates two examples of the lighting unification results. We can see that the lighting effects of the two actors are inconsistent in their original footages. Such inconsistency is undesirable when the actors are expected to interact within the same scene. By applying our model, we can relight them into a common environment with consistent lighting effect. 

\section{Conclusion \& Discussions}
In this work, we have presented a diffusion-based framework for portrait video relighting that achieves high-fidelity photorealism while maintaining temporal and lighting consistency. Central to our approach is a high-quality hybrid dataset comprising both real-world captures and high-fidelity renders, paired with ground-truth environmental illumination maps and per-frame flat-lit albedo references. We construct the Pixel Cube, a cube-shaped LED stage for fast and high-quality video relighting acquisition.
The breadth of our training dataset\textemdash including diverse lighting conditions, subject appearances, and complex facial dynamics\textemdash provides a solid foundation for our relighting model, ensuring strong generalization to novel subjects and environments. Our results on in-the-wild videos demonstrate that the proposed method establishes a new state-of-the-art in identity preservation, lighting harmony, and temporal stability for video portraiture.

\paragraph{Limitations and Future Directions.} 
Despite the advances presented, our system has several limitations for future research. First, the Pixel Cube is constrained by the maximum hardware brightness of the LED panels. This limits the system’s ability to emulate environments with extreme dynamic ranges or high-contrast directional sources, such as direct sunlight during sunrise or sunset. 
Second, the current synchronized acquisition sequence introduces an $8\,\text{ms}$ latency between frames, which can lead to artifacts during extremely rapid subject motion. 
Potential future work could involve interleaving reference flat-lit frames at a lower frequency to further minimize temporal displacement. 
Finally, while our overlapping inference strategy facilitates long-video generation, it remains susceptible to error accumulation over extended durations. Refining the noise-scheduling mechanism or incorporating global temporal constraints could further mitigate potential drift in long video sequences.

\begin{acks}
Yufan Zhang and Jinwei Ye were supported in part by NSF CAREER Award 2238141.
We would like to thank Xinyuan Li, Yanchen Liu, and Minghui Zhao for their valuable suggestions on hardware engineering, Journey Entertainment for providing video materials for our experiments, and anonymous reviewers for their constructive suggestions.
\end{acks}

\bibliographystyle{ACM-Reference-Format}
\bibliography{relight}


\end{document}